\documentclass{article}

\usepackage[final]{corl_2023}

\usepackage{multicol}
\usepackage{wrapfig}
\usepackage{times}
\usepackage{amsmath}
\usepackage{amssymb}
\usepackage{siunitx}
\usepackage{gensymb}
\usepackage{booktabs}
\usepackage{multirow}
\usepackage{authblk}
\usepackage{pifont}
\usepackage{soul}
\usepackage[symbol]{footmisc}
\usepackage{todonotes}
\usepackage{subcaption}
\usepackage{xcolor}
\usepackage{adjustbox}
\usepackage{pdflscape}
\usepackage{bbm}
\usepackage{hyperref}
\usepackage{soul}
\usepackage{listings} %

\definecolor{codegreen}{rgb}{0,0.6,0}
\definecolor{codegray}{rgb}{0.5,0.5,0.5}
\definecolor{codepurple}{rgb}{0.58,0,0.82}
\definecolor{backcolour}{rgb}{0.95,0.95,0.92}
\hypersetup{
   colorlinks=true,
   linkcolor=blue,
   urlcolor=blue,
}
\usepackage[moderate]{savetrees}
\usepackage{enumitem}
\setlist[itemize]{leftmargin=2em}
\setlist[enumerate]{leftmargin=2em}

\definecolor{mygreen}{rgb}{0, 0.5, 0}
\definecolor{mygrey}{rgb}{0.6, 0.6, 0.6}
\newcommand{\task}[1]{\textbf{\footnotesize{\texttt{#1}}}}
\newcommand{\cmark}{\textcolor{green}{\ding{51}}}%
\newcommand{\xmark}{\textcolor{red}{\ding{55}}}%
{\list{}{\leftmargin=0.3in\rightmargin=0.3in}\item[]}%
{\endlist}

\newcommand{\mypara}[1]{\paragraph{#1}}

\newcommand{\ie}{\textit{i}.\textit{e}., }
\newcommand{\eg}{\textit{e}.\textit{g}. }

\definecolor{mygreen}{rgb}{0.78, 1.0, 0.816}
\definecolor{mygreen2}{rgb}{0.235,0.706,0.294}
\definecolor{myred}{rgb}{0.902,0.098,0.294}
\definecolor{myblue}{rgb}{0.263, 0.388, 0.847}

\definecolor{plt_red}{rgb}{0.8588,0.1568,0.1568}
\definecolor{plt_green}{rgb}{0.1725,0.6724,0.1725}
\definecolor{plt_purple}{rgb}{0.34901960784,0.29411764705,0.56862745098}

\DeclareRobustCommand{\hlgreen}[1]{{\sethlcolor{mygreen}\hl{#1}}}

\definecolor{light-gray}{rgb}{0.8, 0.8, 0.8}
\definecolor{comment-green}{rgb}{0.435, 0.576, 0.106}
\definecolor{prompt-blue}{HTML}{2596be}
\definecolor{code-function}{HTML}{379fbe}
\definecolor{code-function}{HTML}{693da8}  %
\definecolor{code-syntax}{HTML}{0060b1}
\definecolor{code-constant}{HTML}{d86001}
\definecolor{prompt-gray}{HTML}{a7a7a7}
\definecolor{highlight}{HTML}{f8f9cb}
\definecolor{highlight}{HTML}{e3eeff}  %
\definecolor{code-perception}{HTML}{2ecc71}
\definecolor{code-control}{HTML}{ff9900}
\definecolor{code-undefined}{HTML}{ff0000}
\renewcommand\fbox{\fcolorbox{light-gray}{white}}

\newcommand{\hlcode}[1]{\colorbox{highlight}{#1}}

\NewDocumentCommand{\code}{v}{%
   \texttt{\small{\textcolor{code-syntax}{#1}}}%
}

\newcommand{\query}[1]{\textcolor{comment-green}{#1}}

\begingroup\lccode`~=`\_ \lowercase{\endgroup\let~}\_
\catcode`\_=12

\title{Scaling Up and Distilling Down: \\ { Language-Guided Robot Skill Acquisition}
}

\author[1]{Huy Ha}
\author[2]{Pete Florence}
\author[1]{Shuran Song}
\affil[1]{Columbia University}
\affil[2]{Google DeepMind}

\begin{document}

\maketitle

\vspace{-10mm}
\begin{abstract}
   We present a framework for robot skill acquisition, which 1) efficiently scale up data generation of language-labelled robot data and 2) effectively distills this data down into a robust multi-task language-conditioned visuo-motor policy.
   For (1), we use a large language model (LLM) to guide high-level planning, and sampling-based robot planners (\eg motion or grasp samplers) for generating diverse and rich manipulation trajectories.
   To robustify this data-collection process, the LLM also infers a code-snippet for the success condition of each task, simultaneously enabling the data-collection process to detect failure and retry as well as the automatic labeling of trajectories with success/failure.
   For (2), we extend the diffusion policy single-task behavior-cloning approach to  multi-task settings with language conditioning.
   Finally, we propose a new multi-task benchmark with 18 tasks across five domains to test long-horizon behavior, common-sense reasoning, tool-use, and intuitive physics.
   We find that our distilled policy successfully learned the robust retrying behavior in its data collection procedure, while improving absolute success rates by $33.2\%$ on average across five domains.
   All code, data, and qualitative policy results are available at \website.
\end{abstract}

\begin{figure}[h]
    \centering
    \includegraphics[width=\linewidth]{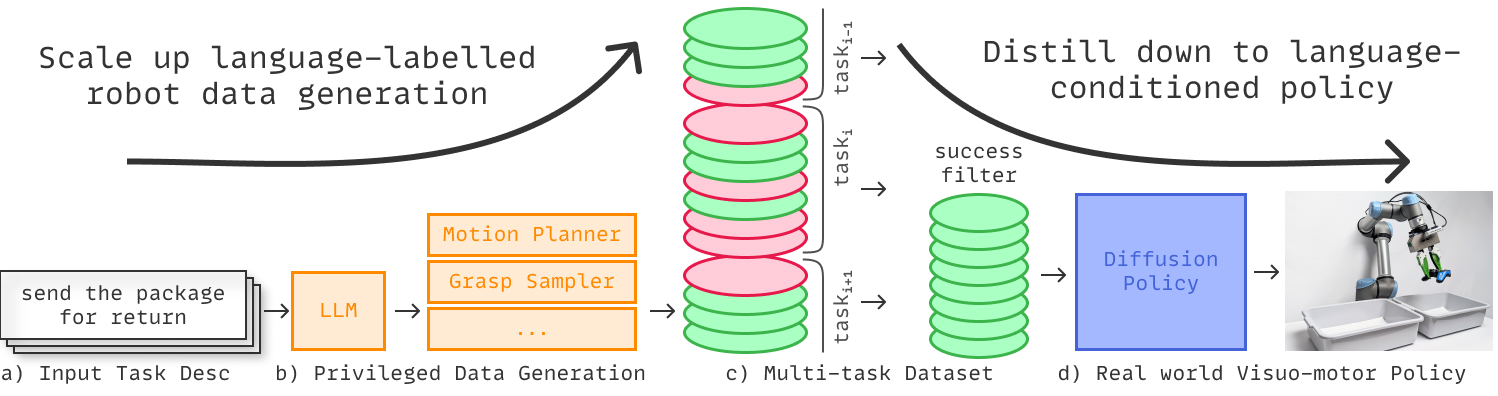}
    \caption{ \footnotesize
        \textbf{Language-guided Skill Acquisition} enables scalable robot learning.
        In the data generation stage, a LLM takes as input task descriptions~(a) and uses sampling-based robotic planners and privileged simulation information~(b) to perform task-directed exploration.
        This enables the scaling up of language and task-success labeled dataset generation~(c).
        In the second stage, the dataset is filtered for success and distilled down into a closed-loop language-conditioned visuomotor policy for real world deployment~(d).
    }
    \vspace{-3mm}
    \label{fig:teaser}
\end{figure}

\vspace{-2mm}
\section{Introduction}
\vspace{-2mm}
How can we scalably acquire robust, reusable, real-world manipulation skills? This question has been the driving force behind extensive research in robot learning.
Attempts in the field have focused on two primary aspects:
First, how to \textbf{scale up the data collection} for a diverse range of manipulation skills, which involves efforts such as improving the hardware~\cite{song2020grasping,zhao2023learning} and software~\cite{todorov2012mujoco,robomimic2021} which support demonstration collection, utilization of non-robotics datasets~\cite{nair2022r3m,grauman2022ego4d}, or trial-and-error explorations~\cite{fu2020d4rl}.
The second aspect of this question concerns \textbf{effective learning} from the collected data, which delves into
exploring effective action representations~\cite{wu2020spatial,shridhar2022peract,shridhar2021cliport} and policy formulations~\cite{florence2021implicit,chi2023diffusionpolicy} that can robustly model the training data and generalize to novel scenarios.

This paper proposes a new framework that provides a comprehensive solution for both aspects by leveraging language guidance, while using no expert demonstrations or reward specification/engineering.
We contribute two key components with our framework:

\begin{itemize}[leftmargin=3mm]
    \vspace{-2mm}
    \item  \textbf{Scaling Up Language-Guided Data Generation: }
          Our data-collection policy is a large language model (LLM) which has access to a suite of 6DoF exploration primitives (\ie sampling-based robot planners and utilities).
          Given an input task description, this policy first \textbf{simplifies} the task by recursively decomposing it into subtasks, resulting in a hierarchical plan (\ie task tree).
          Next, this plan is \textbf{grounded} into a sequence of 6DoF exploration primitives, which generates diverse robot trajectories for the task.
          Finally, the data collection policy \textbf{verifies} the trajectories' success with an inferred success function and \textbf{retries} the task until it succeeds.
          This verify \& retry step not only improves the data-collection policy's success, but also adds robot experience on how to recover from failure, an important trait for downstream policy distillation.
          This data generation approach is scalable, enabling significantly more efficient autonomous task-directed exploration than unguided alternatives (\ie reinforcement learning) while not being limited by the lack of low-level understanding of the LLM-only solution.
    \item  \textbf{Distilling Down to Language-Conditioned Visuomotor Policy:}
          We distill these robot experiences into a visuo-linguo-motor policy that infers control sequences from visual observations and a natural language task description.
          To enable effective learning of high entropy, diverse robot trajectories, we extend the diffusion policy~\cite{chi2023diffusionpolicy} to handle language-based conditioning for multi-task learning.
          This allows the learned policy to be reused and recomposed through language-based planners.
          We found that our distilled policy successfully learned the robust retrying behavior from its data collection policy, while improving upon its absolute success rate across five domains by $33.2\%$.
          Further, we demonstrate that our policy directly transfers to the real-world without fine-tuning using domain randomization.
          \vspace{-2mm}
\end{itemize}

Our framework combines these two components to get the best of both worlds -- leverage LLM's common-sense reasoning abilities for efficient exploration while learning robust and re-usable 6DoF skills for real-world deployment.
In summary, the key contribution of this paper is a new framework for visuo-linguo-motor policy learning that is enabled by three novel components:
\begin{itemize}[leftmargin=3mm]
    \vspace{-2mm}
    \item A new language-guided data collection framework that combines language-based task planner with 6DoF robot utilities (\eg motion planning, grasp sampling).
    \item New formulation of diffusion-based policy that effectively learns multi-task language-conditioned closed-loop control policies.
    \item In addition to our algorithmic contributions, we also contribute a new multi-task benchmark that includes 18 tasks across five domains, requiring long-horizon ($\approx 800$ control cycles), common sense, tool-use, and intuitive physics understanding -- capabilities lacking in existing manipulation benchmarks.
          \vspace{-2mm}
\end{itemize}

\vspace{-3mm}
\section{Related Works} \vspace{-3mm}
\label{sec:related_works}

\mypara{Scaling visuo-linguo-motor data.}
In learning vision-and-language-conditioned motor policies for real-world deployment~\cite{lynch2020language,stepputtis2020language,jang22a,shridhar2021cliport,shridhar2022peract,lynch2022interactive,mees2022matters,brohan2022rt}, one of the most important questions is how to scale up ``robot-complete data'' -- data that has robot sensory inputs (\eg vision), action labels (\eg target end-effector \& gripper commands), and task labels (\eg language description, success).
The most prevalent paradigm is to use humans to annotate both actions (\eg teleoperation) and language~\cite{lynch2020language,stepputtis2020language,jang22a,shridhar2021cliport,shridhar2022peract,lynch2022interactive,mees2022matters,brohan2022rt}.
When providing action labels, humans can either provide task-specific~\cite{jang22a,shridhar2021cliport,shridhar2022peract,brohan2022rt}, or task-agnostic (``play'') data~\cite{lynch2020language,stepputtis2020language,mees2022calvin,lynch2022interactive}.
A primary limitation, however, is that data scalability is human-limited.

Other prior works have proposed strategies to enable more-autonomously-scalable data.
To scale language annotation, prior works study using visual-language models~\cite{xiao2022robotic,zhang2023sprint}, or procedurally post-hoc provided in simulation~\cite{mees2022calvin}.
To scale action labels, methods study how to use {\em{autonomous sub-optimal policies}} from random~\cite{fu2020d4rl} to learned~\cite{nair2022learning} policies.
Human egocentric videos~\cite{goyal2017something,grauman2022ego4d,damen2018scaling} has also been shown to be relevant to robot learning~\cite{chen2021learning,nair2022r3m}, but \emph{is not robot-complete} (lacks action labels), and requires cross-embodiment transfer.
Towards unsupervised exploration, prior works have also investigated evolving environments~\cite{wang2019paired,jiang2021replay} and embodiments~\cite{mouret2015illuminating}, automatic task generation~\cite{fang2022active}, leveraging language guidance~\cite{du2023guiding,mirchandani2021ella} and world-model error~\cite{mendonca2023alan}, but have not been demonstrated to scale to 6 DoF robotic skill learning.
While these approaches reduce human efforts, they are still limited in optimality, generality, and/or completeness of robot data labels.

Another option for the autonomous data collection policy is to use a model-based policy, \eg task and motion planning (TAMP)~\cite{garrett2021integrated}.
Our approach extends such methods in terms of flexibility and task generality by leveraging LLM's common-sense knowledge.
However, in contrast to recent works which use LLMs as the \emph{final} policy~\cite{huang2022language,ahn2022can,huanginner,codeaspolicies2022,driess2023palm,lin2023text2motion,singh2023progprompt}, we use the LLM-based planner as a suboptimal \emph{data-collection} policy.
We then distill only successful trajectories into an observable-information~\cite{agarwal2022legged,seita2019deep,miki2022learning} policy, allowing the distilled policy to improve upon its LLM data collection policy's performance.

\mypara{Policy Representations and Multi-task Policy Distillation.} One primary question in visuo-motor learning~\cite{levine2016end} has been how to represent the policy for effective learning, i.e. to enable high precision, multi-modal robot behavior~\cite{florence2021implicit,hausman2017multi,shafiullah2022behavior,chi2023diffusionpolicy,zhao2023learning}.  Another related question has been how to best train multi-task policies \cite{yu2020meta,kalashnikov2021mt}, including those conditioned on language~\cite{lynch2020language,jang22a,shridhar2021cliport,shridhar2022peract,lynch2022interactive,brohan2022rt}.  Our work presents the novel formulation of bringing diffusion-based \cite{sohl2015deep,ho2020denoising} policies~\cite{chi2023diffusionpolicy} into the language-conditioned~\cite{saharia2022photorealistic,rombach2022high} visuomotor domain.   Additionally, prior works in multi-task language-conditioning typically focus on cloning policies from experts, meanwhile we study distilling data from a success-filtered suboptimal policy.  Success-filtering \cite{florence2021implicit,chen2021decision} can be viewed as the simplest form of offline RL \cite{levine2020offline}.

\begin{figure}[t]
   \centering
   \includegraphics[width=0.98\linewidth]{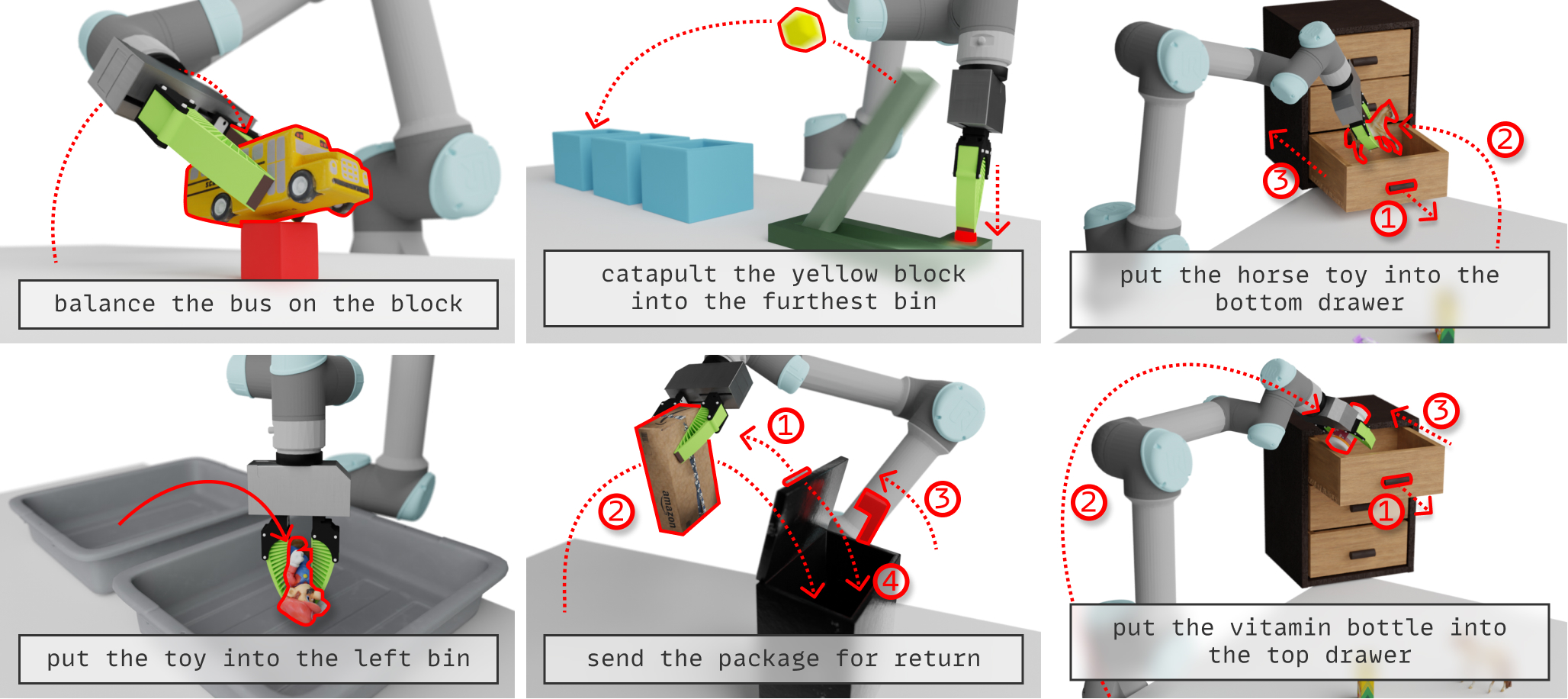} \vspace{-2mm}
   \caption{ \footnotesize
      \textbf{Benchmark.} We validate our approach on a new multi-task benchmark addressing challenging long-horizon tasks (\ie 800 control cycles) requiring language understanding (e.g., put [object] to [top] drawer), common sense knowledge (e.g., send a package for return requires raising the mailbox flag), tool-use (e.g., catapult), and intuitive physics (e.g., balance the bus). The tasks are best viewed on our \website.
   }\vspace{-5mm}
   \label{fig:tasks_panel}
\end{figure}
\vspace{-2mm}
\section{Approach} \vspace{-2mm}

We propose a new framework for robot learning that performs automatic data collection and policy learning from only a task description. Our design is grounded on four key observations:
\vspace{-2mm}
\begin{itemize}[leftmargin=3mm]
   \item We recognize the importance of random exploration in reinforcement learning, but aim to not be constrained by its inefficiency for long-horizon, sparse reward tasks.
   \item We acknowledge the usefulness of LLM's common-sense and zero-shot capabilities, but believe language \emph{is not by itself} the ideal representation for robust, rich, and precise robotic manipulation.
   \item We are inspired by the effectiveness of robotic planning methods, e.g. TAMP,
         but wish to be flexible to novel tasks and domains %
         and non-reliant on ground truth state %
         during policy inference.
   \item We aim to achieve the simplicity and effectiveness of behavior cloning in distilling collected robot experience into a policy for real-world deployment, while side-stepping the requirement for costly human demonstrations or play data collection.
         \vspace{-2mm}
\end{itemize}
Using no human demonstration or manually specified reward, our framework combines the strengths of these four areas into a unified framework for both efficient task-directed exploration and multi-task visuo-linguo-motor policy learning.

\textbf{Method Overview.}
In the data generation phase, we use an LLM to recursively decompose (\S\ref{sec:method:datagen:planning}) tasks into a hierachical plan (\ie task tree) for exploration and ground the plan into sampling-based robot utilities and motion primitives (\S\ref{sec:method:datagen:grounding}).
Next, the LLM infers success-detection functions for each task in the plan (\S\ref{sec:method:datagen:verifying}), providing success-labeling.
This autonomous data generation process outputs a replay buffer of task-directed exploration experience, labeled with language descriptions and success labels.
In the training phase (\S \ref{sec:method:distill}), we filter this data for success according to the LLM inferred success condition and distill it into a multi-task vision-and-language-conditioned diffusion policy~\cite{chi2023diffusionpolicy}.

\begin{figure}[t]
   \centering
   \includegraphics[width=\linewidth]{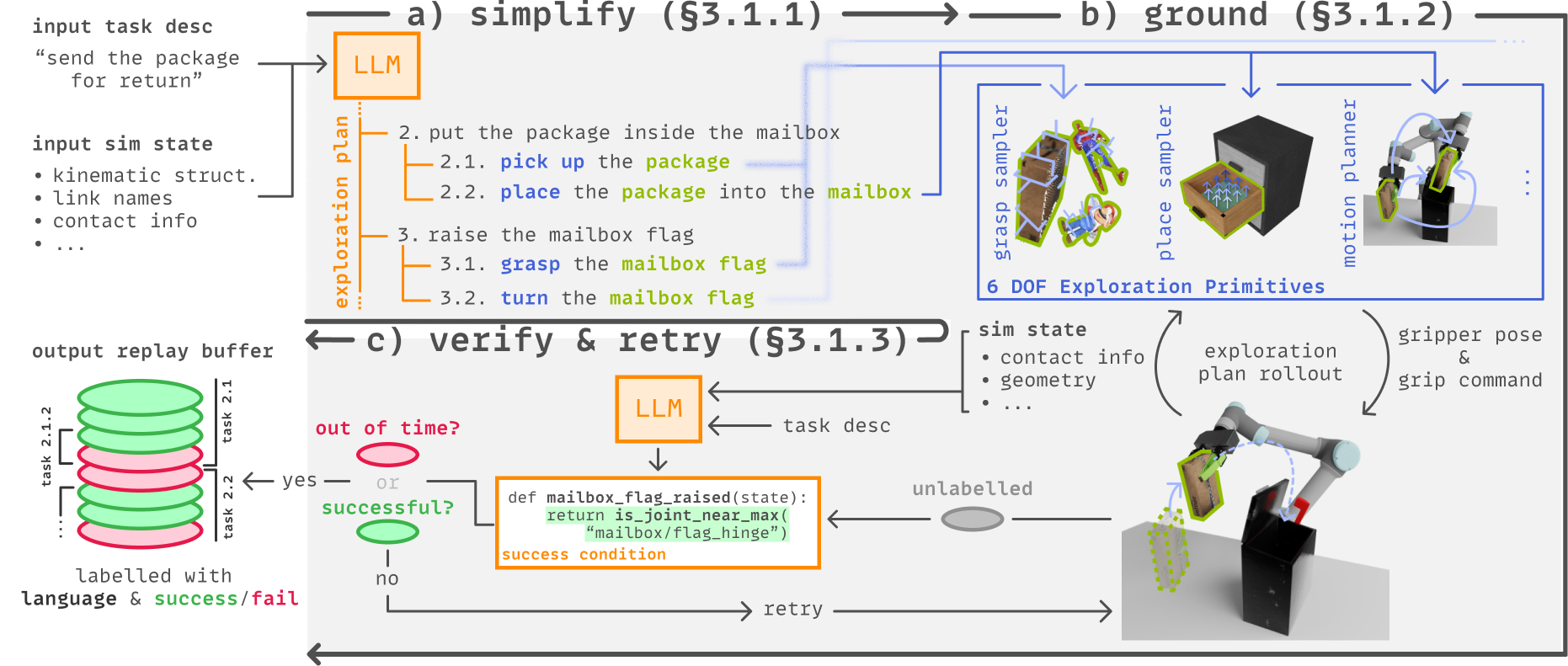}
   \caption{ \footnotesize
      \textbf{Language-Driven Robot Data Generation} takes as input the task description and simulation state, and outputs a replay buffer, labelled with language descriptions and success.
      It starts by using an LLM to simplify tasks recursively (a) until the task involves only one object, resulting in a hierarchical exploration plan.
      Next, the plan is grounded (b) into a sequence of \textcolor{myblue}{6 DOF exploration primitives} (\eg grasp samplers, motion planners, etc.) and rolled out in simulation to give an unlabelled robot trajectory.
      Finally, an LLM infers a \hlgreen{success function code-snippet}, and uses it to verify (c) and label it with \textcolor{mygreen2}{succeeded} or \textcolor{myred}{failed}.
      If the trajectory failed, the LLM retries the exploration plan with a different random seed (\eg a different grasp pose from the grasp sampler).
      If the robot succeeds or run out of time, the labeled trajectory is returned.
   }
   \vspace{-5mm}
   \label{fig:method:datagen}
\end{figure}
\vspace{-2mm}

\subsection{Simplify: Task Planning and Decomposition}\vspace{-2mm}
\label{sec:method:datagen:planning}

Given a task description, the first step is to generate a high-level task plan. To improve the flexibility to work with any tasks and 3D assets, we opted for an LLM-based planner to leverage their common-sense and zero-shot reasoning skills. Unlike classical TAMP planners, our framework does not require  domain-specific engineering and transition function design to work with new tasks. %

Concretely,  our recursive LLM planner takes as input the task description, the simulation state, and outputs a plan in the form of a task tree (Fig.~\ref{fig:method:datagen}a).
To do so, the LLM first checks whether the task description involves the robot interacting with multiple or only one object.
For instance, ``move the package into the mailbox'' involves opening the mailbox before picking up the package and putting the mailbox in, and should be considered a multi-object task.
Meanwhile, ``with the mailbox opened, move the package into the mailbox'' should be a single-object task.
For the base case of single-object tasks, we prompt the LLM to which object part name to to interact.
For the case of multi-object tasks, we prompt the LLM to decompose the task into subtasks, and recurse down each subtask.

\vspace{-2mm}
\subsection{Ground: Compiling a Plan into Robot Utilities}\vspace{-2mm}
\label{sec:method:datagen:grounding}
With the generated task tree \S \ref{sec:method:datagen:planning}, the next step is to ground the high-level plan into physical actions. Here, the choice of the \textit{low-level robot API} critically defines the system's capability and, therefore, becomes a key differentiating factor between different systems.
In principle,  there are three desired properties we want to see in the action space design:
\begin{itemize}[leftmargin=3mm]
   \vspace{-2mm}
   \item \textbf{Flexibility.} Planar actions~\cite{codeaspolicies2022,shridhar2021cliport} aren't flexible enough to manipulate prismatic and revolute joints.
   \item  \textbf{Scalable.} Namely, actions should not require human demonstrations to acquire~\cite{jang22a,ahn2022can,lynch2020language,stepputtis2020language,shridhar2021cliport,shridhar2022peract,lynch2022interactive}.
   \item \textbf{Language-friendly.} While joint sequences can encode any action, it is not language-friendly.
         \vspace{-2mm}
\end{itemize}

We propose to ground the LLM's plan with API calls into a set of robot utility functions, which include a sampling-based motion planner, a geometry-based grasp and placement sampler, and motion primitives for articulated manipulation.
We refer to these utilities as 6 DOF Exploration Primitives (Fig~\ref{fig:method:datagen}b) because, by virtue of being \emph{pseudo-random}, the sampling-based utilities generate \emph{diverse} robot trajectories, enabling effective exploration for rich 6 DoF manipulation settings.
For instance, our grasp and placement samplers samples uniformly amongst all points in the object part's point cloud to find good grasps and placements poses, respectively, which are used as input into a rapidly-exploring random trees~\cite{lavalle1998rapidly} motion planner that samples uniformly in joint space.
This results in diverse grasps, placements, and motion trajectories connecting grasps and placements.

For each leaf node in the inferred task tree (\S~\ref{sec:method:datagen:planning}), the grounding process takes as input the node's task description (\eg ``open the mailbox''), its associated object part name (\eg ``mailbox lid''), and the simulation state, and outputs a sequence of 6 DoF Exploration Primitive API calls.
Using the object part name, we can parse the object's kinematic structure from the simulation state and handle articulated and non-articulated (\ie rigid, deformable) objects separately.
For non-articulated objects, the LLM is prompted to choose the pick \& place object names, used to sample grasp and placement pose candidates.
For articulated objects (with either revolute or prismatic joints), the leaf node's associated object part name is used to sample a grasp candidate followed by a rotation or translation primitive conditioned on its joint parameters (i.e., joint type, axis, and origin).

\textbf{Exploration Plan Rollout.}
Each node in the exploration plan is grounded only when it is being executed, where the order of execution follows a pre-order tree traversal.
By keeping track of the subtask's state, sub-segments of robot trajectory can be labelled with the subtask's description, thereby providing \textbf{dense and automatic text labels} for the trajectory.
For instance, all actions taken during the inferred subtask ``open the mailbox'' can be labeled with both the subtask's description ``open the mailbox'' and the root task description ``move the package into the mailbox''.

Since grounding happens only when a task node is visited, each node's grounding process is independent of the other leaf nodes, depending only on the simulation state when it is evaluated.
While this simplifies planning significantly, it also means that failed execution can occur.
For instance, a grasp candidate may render all placement candidates infeasible. %

\vspace{-2mm}
\subsection{Verify \& Retry: Robustifying the Data Collection Policy}\vspace{-2mm}
\label{sec:method:datagen:verifying}

Recall, the planning and grounding step can fail, especially when we consider long-horizon tasks.
To address this, we propose a verify \& retry (Fig.~\ref{fig:method:datagen}c) scheme, which uses environment feedback to detect failed execution.

\textbf{Verify.} For each task, the LLM infers \textbf{a success function code snippet} given the task description, simulation state, and API functions to for query simulation state (e.g., checking contact or joint values, etc).
This amounts to prompting the LLM to complete a task success function definition that outputs a boolean value, indicating task success.
For instance, given the task ``raise the mailbox flag'', the LLM's inferred code snippet should check whether the mailbox's flag hinge is raised (Fig.~\ref{fig:method:datagen}c, highlighted green).

\textbf{Retry.} When a trajectory is labeled failed, the robot retries the same sequence of robot utilities with a different random seed (\ie for the sampling-based robotic utilities) without resetting the simulation state until the task succeeds.
For instance, in the bus balance task (Fig.~\ref{fig:tasks_panel}, top left), the robot would repeatedly try different grasp and place candidates until the bus is balanced.
In the tree traversal process \S~\ref{sec:method:datagen:grounding}, nodes only yield execution to its parent task when the node's inferred success condition returns true.
This design not only leads to higher success rates in data generation but also provides useful demonstrations on \textbf{how to recover from failure}.
In the output replay buffer, the only failed trajectories are ones which timed-out or led to invalid states (\eg object dropped on the floor).

\begin{wrapfigure}{r}{0.5\textwidth}
   \centering
   \vspace{-12mm}
   \includegraphics[width=\linewidth]{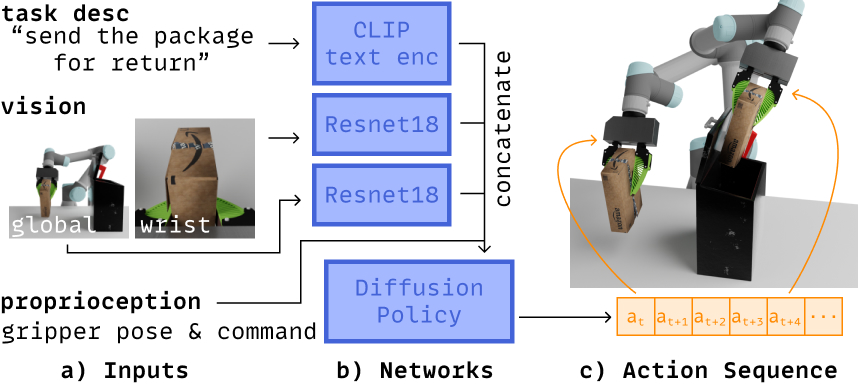}
   \caption{ \footnotesize
      \textbf{Language-Conditioned Policy Distillation}. The policy takes as input a task description, two RGB camera views, and gripper proprioception data, and outputs a sequence of gripper poses and closing command.
   }
   \label{fig:method:distillation}
   \vspace{-4mm}
\end{wrapfigure}

\vspace{-2mm}

\subsection{Language-conditioned Policy Distillation}
\label{sec:method:distill}
\vspace{-2mm}

We extend diffusion policy~\cite{chi2023diffusionpolicy}, a state-of-the-art approach for single-task behavior cloning, to the multi-task domain by adding language-conditioning.
This policy takes as input a task description CLIP~\cite{radford2021learning} feature, proprioception history, and visual observations, and outputs a sequence of end effector control commands.
Following Robomimic~\cite{robomimic2021}'s findings, we use a wrist-mounted view in addition to a global (workspace) view to help with tasks requiring precise manipulation.
We use their ResNet18-based~\cite{he2016deep} vision encoders, one for each view.
We found that using only the latest visual observation along with the full observation horizon of proprioception maintains the policy's high performance while reducing training time.
When used in conjunction with the DDIM~\cite{songdenoising} noise scheduler, we found that we could use a $10\times$ shorter diffusion process at inference (5 timesteps at inference, 50 timesteps at training) while retaining a comparable performance.
Quantitatively, when using a 10 dimensional action space\footnote{3 for position, 6 for rotation using the upper rows of the rotation matrix, and a gripper close command}, our policy can be run at $\approx 35Hz$ on an NVIDIA RTX3080.

\vspace{-2mm}
\section{Evaluation}

\vspace{-2mm}
\label{sec:experiment}

\begin{wraptable}{r}{0.4\linewidth}
    \centering
    \vspace{-18mm}
    \setlength{\tabcolsep}{0.02cm}
    \scriptsize
    \centering{
        \begin{tabular}{lcccccc}
            \toprule
            Domain    & \begin{tabular}[c]{@{}c@{}}Complex\\geometry\end{tabular} & \begin{tabular}[c]{@{}c@{}}Artic-\\ulation\end{tabular} & \begin{tabular}[c]{@{}c@{}}Common\\sense\end{tabular} & \begin{tabular}[c]{@{}c@{}}Tool\\use\end{tabular} & \begin{tabular}[c]{@{}c@{}}Multi-\\task\end{tabular} & \begin{tabular}[c]{@{}c@{}}Long\\horizon\end{tabular} \\
            \midrule
            Balance   & \xmark                    & \xmark                    & \xmark                    & \xmark                    & \xmark                    & \xmark                    \\
            Catapult  & \xmark                    & \cmark                    & \cmark                    & \cmark                    & \cmark                    & \xmark                    \\
            Transport & \cmark                    & \xmark                    & \xmark                    & \xmark                    & \xmark                    & \xmark                    \\
            Mailbox   & \xmark                    & \cmark                    & \cmark                    & \xmark                    & \xmark                    & \cmark                    \\
            Drawer    & \cmark                    & \cmark                    & \xmark                    & \xmark                    & \cmark                    & \cmark                    \\
            \bottomrule
        \end{tabular} \vspace{-2mm}
        \caption{
            \textbf{Benchmark Suite.}
        }
        \vspace{-6mm}
        \label{tab:benchmark}
    }
\end{wraptable}

Our experiments try to validate two questions:
1) Can our data generation approach efficiently perform task-directed exploration?
2) Can our policy learning approach effectively distill a multi-modal, multi-task dataset into a generalizable and robust visuo-linguo-motor policy?

\textbf{Our Benchmark} contains 18 tasks across 5 domains (Fig.~\ref{fig:tasks_panel} Tab. \ref{tab:benchmark}), with the following properties:

\vspace{-2mm}
\begin{itemize}[leftmargin=3mm]
      \item  \textbf{6DoF \& articulated manipulation}, for deadling with complex object geometry and articulation.
      \item \textbf{Geometry Generalization.} In our bin transport domain, the robot must generalize its bin transport skill to unseen object instances, with novel shapes, sizes, and colors.
      \item \textbf{Intuitive physics.}
            Robots should understand the physical properties of the world and use this knowledge to perform tasks.
            In the bus balance domain, the robot needs to learn the precise grasping and placement to balance a large bus toy on a small block.
            In the catapult domain, where the block is placed along a catapult arm determines how far the block will be launched, and, thus, which bin (if any) the block will land in.

      \item \textbf{Common-sense reasoning \& Tool-use.}
            Natural language task description is user-friendly but often under-specifies the task.
            Common-sense can help to fill in the gaps.
            In the mailbox domain, given the task ``send the package for return'', the robot should understand that it not only needs put the package inside, but also raise the mailbox flag to indicate that the package is ready for pickup.
            In the catapult domain, the robot needs to understand that pressing the catapult's button will activate the catapult, and that the block needs to be placed on the catapult arm to be launched.

      \item \textbf{Multi-task conditioning.}
            Given the same visual observations but different task description, the robot should perform different and task-relevant actions. The catapult domain has 3 tasks for three target bins, and the drawer domain has 12 tasks.

      \item \textbf{Long horizon behaviour.}
            Our longest horizon domain, mailbox, takes at least 4 subtasks to complete (open the mailbox, put the package in the mailbox while its opened, close the mailbox, then raise the mailbox flag) which can require up to 800 control cycles.
            In the drawer domain, the robot needs to open the drawer, move the object into the drawer, then close it, which takes about 300 control cycles.
            \vspace{-2mm}
\end{itemize}

The benchmark is built on top of the MuJoCo~\cite{todorov2012mujoco} simulator, using assets from the Google Scanned dataset~\cite{downs2022scannedobjects,zakka2022scannedobjectsmujoco}. We use a table-top manipulation set-up with a 6DoF robot arm.
The task success in evaluation is a manually designed function, instead of LLM generated function used for data collection.

\begin{figure}[t]
    \centering
    \vspace{-6mm}
    \includegraphics[width=\linewidth]{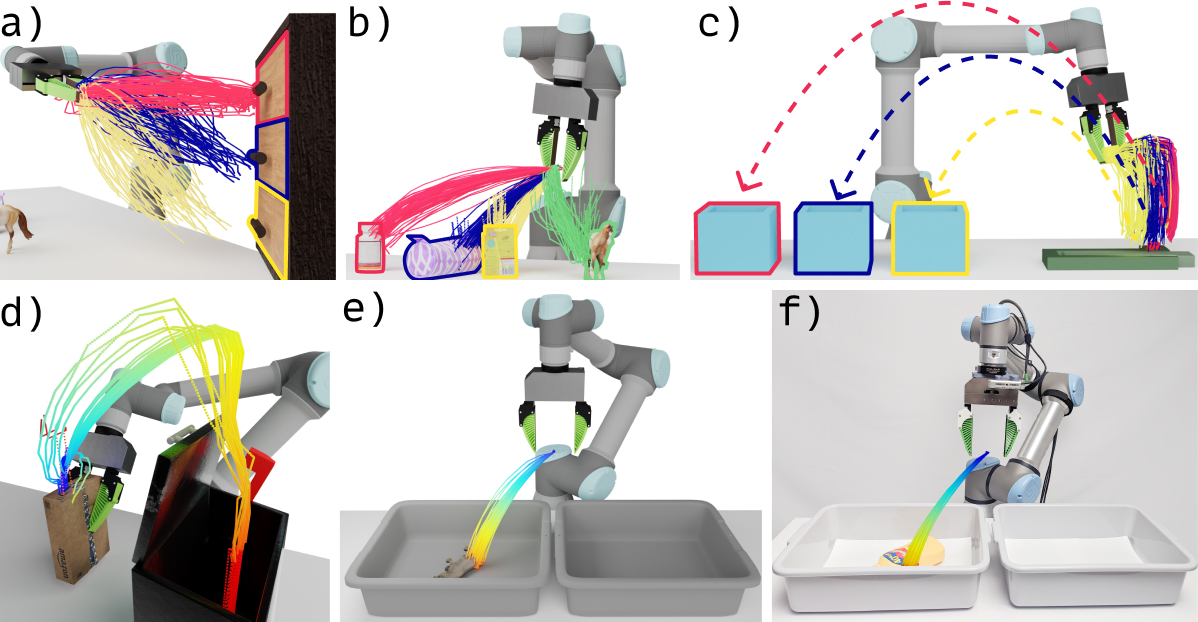}
    \caption{ \footnotesize
        \textbf{High Entropy yet Precise Language-Guided Action Sequences.}
        Running the pseudorandom language-conditioned diffusion process with different seeds on the same observations yields language-consistent (a-c, different colors for different task descriptions), high entropy actions when possible (a-f, object grasping, transports, \& placements) and precise actions when necessary (d, narrow mailbox with large package).
        Further, domain randomization enables a simulation trained policy (e) to generalize to the real world (f).}

    \label{fig:result:retry_from_failure}
    \vspace{-5mm}
\end{figure}

\vspace{-1mm}
\textbf{Metrics.}
We report the success rates (\%) averaged over 200 episodes in Table~\ref{tab:sim}, a task completion efficiency plot in Fig.~\ref{fig:result:efficiency}, and qualitative results in Fig.~\ref{fig:result:retry_from_failure}.
If a domain has multiple tasks then we report the average performance of all tasks.
We also compare different LLMs in Table~\ref{tab:llm} (10 samples per task) and investigate the sources of error in our system for the mailbox domain in Table~\ref{tab:failure_analysis} (200 trials per execution).

\vspace{-1mm}
\textbf{Data Generation Baselines.}
Code-as-Policy~\cite{codeaspolicies2022} is a state-of-the-art approach for using an LLM directly as a robot policy by making state (\eg query present objects) and action primitive API calls to a robot.
Given an LLM-inferred code string, they execute the snippet in an open-loop fashion.
Crucially, in their table top manipulation setting, they assume access to planar action primitives.
Thus, we introduce the following baselines, which build on top of Code-as-Policy and each other as follows:
\vspace{-2mm}
\begin{itemize} [leftmargin=3mm]
    \item \textbf{LLM-as-Policy (2D)}: Similar to code-as-policy using planar pick-and-place, but we use ground truth object segmentation instead of their off-the-shelf object detectors~\cite{kamath2021mdetr,gu2021open}.
    \item \textbf{(+) 6 DOF robot utils}: Builds on top of the previous baseline by adding access to 6 DOF robot utilities for grasping, placement, motion planning, and articulated manipulation.
    \item \textbf{(+) Verify \& Retry}:
          Adding to the previous baselines, this baseline uses the LLM's predicted success condition to label trajectories and retry failed ones.
          Since the robot utilities involve pseudo-random samplers (\eg RRT, grasp sampling), retrying the task means running these samplers again using the pseudo-random state and environment state from where failed trajectory left it.
          Since we use this approach as our data generation policy, it also serves as an ablation of our approach.
          \vspace{-2mm}
\end{itemize}

\textbf{Policy Distillation Ablations.}
We compare against BC-Z~\cite{jang22a}'s single-task policies which does not use FiLM conditioning (used in their bin emptying and door opening tasks).
To understand the effects of our policy learning design decisions in the single-task regime, we fix training time and dataset size (2 days using at least 500 successful trajectories), and provide the following ablations:
\vspace{-2mm}
\begin{itemize} [leftmargin=3mm]
    \item \textbf{Action Generation}:
          Instead of using diffusion processes conditioned on the policy input embedding to decode actions, it is typical use multi-layer perceptrons.
          Following \citeauthor{jang22a}~\cite{jang22a}, we use one \textbf{MLP} with two hidden layers and ReLU activations for end effector position, one for the orientation, and another for gripper command.
          This standard policy architecture is deterministic, and is trained with mean-squared error loss for pose and binary cross entropy loss for gripper command.
    \item \textbf{Action Space}:
          Besides our absolute end effector pose action space, \textbf{Delta-Action} and velocity control spaces is another popular action space choice~\cite{jang22a,robomimic2021,zhang2018deep,florence2019self,mandlekar2020iris}.
          We also ablate BC-Z's execution action horizon (Exec) while keeping their original prediction horizon (Pred).
    \item \textbf{Observation Encoder}: All approaches encode images using a ResNet18~\cite{he2016deep} architecture. Although the original architecture was designed with an average pooling layer, its typical for robotic policies to use a spatial softmax pooling~\cite{levine2016end} layer instead.
    \item \textbf{Data usage}: \textbf{No-Retry} trains on successful trajectories generated from the data generation approach without Verify \& Retry, so it does not observe any recovery behavior.
\end{itemize}

\vspace{-2mm}
\subsection{Data Collection Policy Evaluation}
\vspace{-2mm}
\begin{wraptable}{r}{7.4cm}
      {
            \vspace{-7mm}
            \setlength{\tabcolsep}{0.05cm}
            \scriptsize
            \begin{tabular}{lrrrrrr}
                  \toprule
                  \multirow{2}{*}{Approach} & \multicolumn{3}{c}{Planar} & \multicolumn{2}{c}{6DoF} & \multirow{2}{*}{\scriptsize{Average}}                                              \\\cmidrule(lr){2-4}\cmidrule(lr){5-6}
                                            & \scriptsize{Balance}       & \scriptsize{Catapult}    & \scriptsize{Transport}                & \scriptsize{Mailbox} & \scriptsize{Drawer} \\
                  \midrule
                  LLM-as-Policy (2D)        & 28.0
                                            & \textbf{33.3}
                                            & 21.5
                                            & 0.0
                                            & 0.0
                                            & 27.6
                  \\
                  (+) 6DoF Robot Utils      & 5.5
                                            & 2.5
                                            & 35.0
                                            & 0.0
                                            & 1.3
                                            & 8.8
                  \\
                  (+) Verify \& Retry       & \textbf{45.0}
                                            & 7.3
                                            & \textbf{82.0}
                                            & \textbf{3.0}
                                            & \textbf{31.8}
                                            & \textbf{33.8}
                  \\\midrule
                  Distill No Retry          & 67.5
                                            & 38.5
                                            & 32.5
                                            & 0.0
                                            & 22.7
                                            & 32.2
                  \\
                  Distill Ours              & \textbf{79.0}
                                            & \textbf{58.3}
                                            & \textbf{80.0}
                                            & \textbf{62.0}
                                            & \textbf{55.8}
                                            & \textbf{67.0}
                  \\
                  \midrule
            \end{tabular}
            \vspace{-3mm}
            \caption{
                  \textbf{Success Rates (\%)} for data generation (top) and distillation approaches (bottom) over 200 trials.
            }
            \vspace{-5mm}
            \label{tab:sim}
      }
\end{wraptable}

\textbf{6DoF exploration is critical.}
First, we verify different approach's ability to perform and explore in 6DoF, which is crucial for general manipulation.
When 6DoF exploration is introduced, we first observe a drop in the average success rate for simple tasks that could be accomplished with planar actions (Balance, Transport, Tab. \ref{tab:sim}). However, this ability is critical for exploring complex tasks, providing data to improve upon in the later distilling stage.
In particular, we observed that 6DoF actions are important for grasping diverse objects with complex geometry (Transport, Tab. \ref{tab:sim}), and manipulating articulated objects (Drawer, Mailbox, Tab. \ref{tab:sim}).

\begin{wraptable}{r}{5.0cm}
    {
        \setlength{\tabcolsep}{0.05cm}
        \scriptsize
        \vspace{0mm}
        \begin{tabular}{llrr}
            \toprule
            Subtask                & Planning & Verify & Execution \\\midrule
            Open mailbox           & 100      & 100    & 43.5      \\
            Put package in mailbox & 100      & 100    & 28.5      \\
            Raise mailbox flag     & 100      & 100    & 62.0      \\
            Close mailbox          & 100      & 100    & 94.2      \\
            \midrule
        \end{tabular}
        \vspace{-3mm}
        \caption{
            \textbf{Sources \& Propagation of Error}.
            Accuracy (\%) of planning, verification, and execution success rate (\%) for each mailbox subtask.}
        \label{tab:failure_analysis}
    }
    \vspace{-4mm}
\end{wraptable}

Moreover, 6DoF exploration also helps in \textbf{diversifying} the data collection strategy, which provides the \textbf{possibility to improve upon} in the later distilling stage.
For example in the catapult domain, LLM-as-Policy (2D) is only able to solve one of three possible goals (the closest bin) using a deterministic strategy.
However, it provides no useful data for learning the other two goals, making it a poor data-collection policy.
In contrast, incorporating 6 DOF robot utilities achieves lower but non-zero average success rates in all bins ($16.3\%$, $3.3\%$, and $2.2\%$, full table in appendix), which provide much better exploration data for distillation.

\textbf{Verify \& Retry always helps.}
In the verify \& retry step, the LLM retries all tasks until they are successful.
This simple addition improves performance in all domains, with
$2\times$, $3\times$, $8\times$, and $13\times$ in transport, catapult, balance, and drawer domains.
Without this crucial step, we observe $0.0\%$ success rate in the mailbox domain, underscoring the difficulty of flawlessly executing long sequences of 6 DOF actions, and the importance of recovery after failure.

\begin{wraptable}{r}{4.0cm}
    {
        \vspace{-8mm}
        \setlength{\tabcolsep}{0.11cm}
        \scriptsize
        \begin{tabular}{lrrr}
            \toprule
            Model  & Size          & Planning      & Success \\
            \midrule
            LLAMA2 & 7B            & 42.0
                   & 10.0
            \\
                   & 13B           & 62.0
                   & 48.3
            \\
            \midrule
            GPT3   & 175B          & \textbf{82.0}
                   & \textbf{91.1}
            \\
            \midrule
        \end{tabular}
        \vspace{-2mm}
        \caption{
            \label{tab:llm}
            \textbf{LLM Evaluation}.
        }
    }
    \vspace{-4mm}

\end{wraptable}

\textbf{Language Model Scaling.}
In addition to the final task success, we provide more detailed analysis of planning and success condition inference accuracy in Tab.~\ref{tab:llm}.
We evaluate on the proprietary GPT3~\cite{brown2020language} (175B text-davinci-003) and the open LLAMA2~\cite{touvron2023llama} (7B and 13B).
We found that Llama models struggles in complex planning domains because they do not follow instructions provided in the prompts.
For instance, in the drawer domain, both models fail to account for drawer opening and closing.
However, we observe an upwards trend with respect to Llama model size, with the 13B model outperforming the 7B model by $+20.0\%$ and $+38.3\%$ in planning and success verification accuracy respectively.

\vspace{-2mm}
\subsection{Distilled Policy Evaluation}
\vspace{-2mm}

\textbf{Robustness In, Robustness Out.}
By filtering trajectories with LLM's inferred success condition, distilled policies inherit the robustness of their data collection policies while improving upon success rates ($+23.4\%$ and $+33.2\%$ for no-retry and ours, Tab. \ref{tab:sim}).
Since our distilled policy learned from a robust data collection policy, it also recovers from failures (\eg failed grasps or placements) and continuously retries a task until it succeeds.
Meanwhile, since the no-retry distilled policy learned from a data collection policy which did not retry upon failure, it is sensitive and brittle, leading to $-34.8\%$ lower average success rate across all domains compared to ours (Tab. \ref{tab:sim}).

\begin{wrapfigure}{r}{0.45\textwidth}
    \vspace{-5mm}
    \centering
    \includegraphics[width=\linewidth]{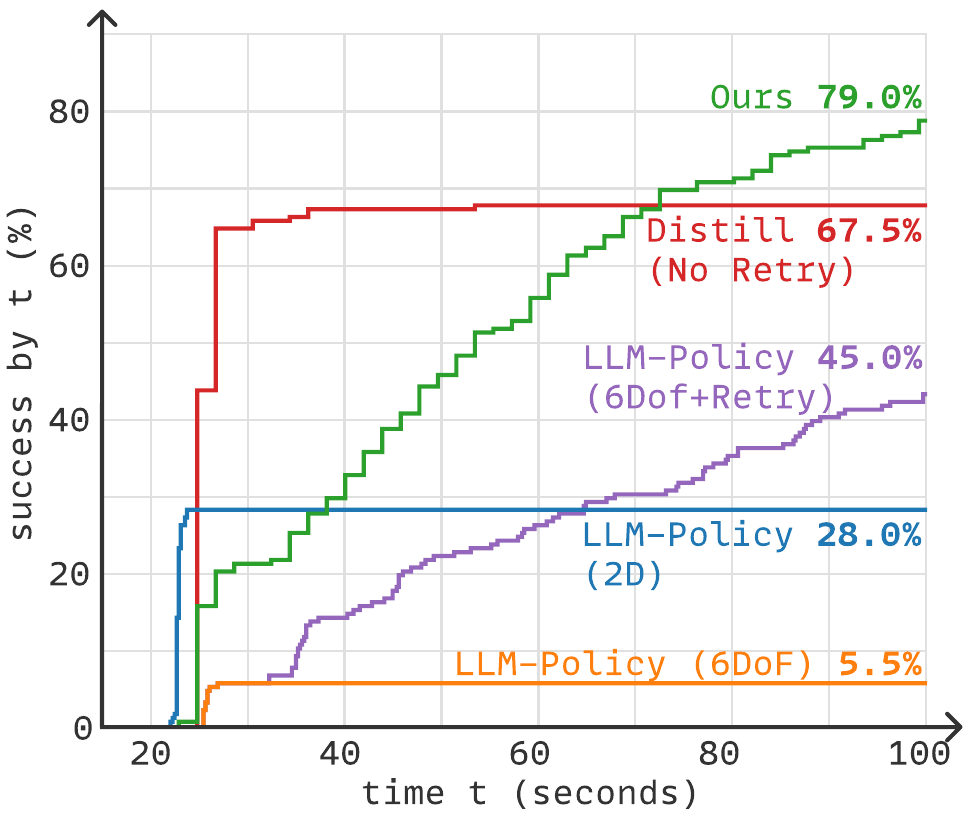}
    \vspace{-6mm}
    \caption{\footnotesize \textbf{Distilled Robustness}.
        \textbf{\textcolor{plt_green}{Our policy}} inherits robust recovery from failure behavior from \textbf{\textcolor{plt_purple}{its data collection policy}}, while improving upon success rate.
    }
    \label{fig:result:efficiency}
    \vspace{-7mm}
\end{wrapfigure}

\textbf{High Performance From Diverse Retry Attempts.}
Plotting how long policies take to solve the balance task (Fig.~\ref{fig:result:efficiency}), we observed that our policy and its data collection policy continuously tries a diverse set of grasps and placements after each failed attempt until it succeeds.
This results in higher success rates as the policy is given more time, and is reflected in their monotonically increasing success rates.
In contrast, baselines plateau after their first grasp/platement attempts.
This highlights the synergy of two design decisions.
First, the verify \& retry step (\S~\ref{sec:method:datagen:verifying}) is crucial for demonstrating retrying behavior, but is by itself \emph{insufficient} if each retrying action is the identical as the previous one.
Instead, opting for a diffusion policy (\S~\ref{sec:method:distill}) for learning from and generating high-entropy, diverse retry attempts (Fig~\ref{fig:result:retry_from_failure}) is also essential for high performance.

\textbf{Policy Learning Baselines.}
We investigate policy learning design decisions on the single-task balance domain, and remove language conditioning.
While BC-Z found spatial softmax hurt their performance and opted for a mean pool, we observed using spatial softmax improved performance by +$5.0$\%.
Further, we found that switching from delta to absolute action spaces improved success rates $+6.5\%$ and $+9.5\%$ when using the MLP action decoder and our diffusion action decoder, respectively, confirming \citeauthor{chi2023diffusionpolicy}~\cite{chi2023diffusionpolicy}'s findings.
Lastly, we find that using our pseudo-random diffusion-based action encoder consistently outperforms a deterministic MLP action mappings, regardless of other design decisions.

\begin{wraptable}{r}{5.7cm}
    {
        \setlength{\tabcolsep}{0.015cm}
        \scriptsize
        \vspace{-4mm}
        \begin{tabular}{lllcclcc}
            \toprule
            \multirow{2}{*}{Method} & \multicolumn{4}{c}{Output} & \multicolumn{2}{c}{Input} & Success                                            \\\cmidrule(lr){2-5}\cmidrule(lr){6-7}
                                    & Generation                 & Rep.                      & Exec    & Pred & Pool    & Proprio & (\%)          \\
            \midrule
            BC-Z                    & FeedForward                & Delta                     & 1       & 10   & Avg     & \xmark  & 0.0           \\
                                    & FeedForward                & Delta                     & 4       & 10   & Avg     & \xmark  & 15.0          \\
                                    & FeedForward                & Delta                     & 8       & 10   & Avg     & \xmark  & 18.5          \\
            \midrule
            Ours
                                    & FeedForward                & Delta                     & 8       & 16   & Spatial & \cmark  & 29.0          \\
                                    & FeedForward                & Abs                       & 8       & 16   & Spatial & \cmark  & 35.5          \\
                                    & Diffusion                  & Delta                     & 8       & 16   & Spatial & \cmark  & 69.5          \\
                                    & Diffusion                  & Abs                       & 8       & 16   & Avg     & \cmark  & 76.5          \\
                                    & Diffusion                  & Abs                       & 8       & 16   & Spatial & \cmark  & \textbf{79.0} \\
            \midrule
        \end{tabular}
        \vspace{-3mm}
        \caption{
            \textbf{Policy Learning Ablations}.
            Action generation using diffusion models~\cite{ho2020denoising} robustly outperforms feed-forward models across other policy design decisions.
        }
        \label{tab:policy}
        \vspace{-6mm}
    }
\end{wraptable}
\textbf{Sim2Real Transfer.}
We evaluated a policy trained on domain randomized synthetic data in a real world transport task with five novel objects (Fig.~\ref{fig:result:retry_from_failure}e).
Averaging across ten episodes per object, our policy achieved 76\% success rate, demonstrating the effectiveness of our approach in Sim2Real transfer.

\vspace{-3mm} \subsection{Limitations} \vspace{-2mm}

By using priviledged simulation state information, the LLM can infer success conditions which uses ground truth contact, joint information, and object poses.
This means our implementation of the data generation phase is limited to simulation environments, and our policy requires sim2real transfer.
Further, Our data generation method relies on existing 3D assets and environments, which presents a further opportunity for scaling up with assets from 3D generative models or procedural generation.
Finally, while our approach's dataset contains text labels and success labels for all subtasks, we have only evaluated its effectiveness in learning the root task.
Learning from all subtasks and growing a robot's set of learned, reusable sub-skills over time to enable compositional generalization is left for future work.

\vspace{-2mm}
\section{Conclusion}
\vspace{-2mm}

We proposed ``Scaling Up and Distilling Down'', a framework that combines the strengths of LLMs, sampling-based planners, and policy learning into a single system that automatically generates, labels, and distills diverse robot-complete exploration experience into a multi-task visuo-linguo-motor policy.
The distilled policy inherits long-horizon behaviour, rich low-level manipulation skills, and robustness from its data collection policy while improving upon performance beyond its training distribution.
We believe that this integrated approach is a step towards putting robotics on the same scaling trend as that of LLM development while not compromising on the rich low-level control.

\acknowledgments{
   We would like to thank Cheng Chi, Zeyi Liu, Samir Yitzhak Gadre, Mengda Xu, Zhenjia Xu, Mandi Zhao and Dominik Bauer for their helpful feedback and fruitful discussions.
   This work was supported in part by Google Research Award, NSF Award \#2143601, and \#2132519. We would like to thank Google for the UR5 robot hardware.  The views and conclusions contained herein are those of the authors and should not be interpreted as necessarily representing the official policies, either expressed or implied, of the sponsors.}
\bibliography{references}

\newpage
\begin{appendix}
   \section{Policy Rollout Visualizations}
\label{sec:supp:policy_rollout_vis}

Our policy's 6DoF manipulation behavior is best visualized through videos.
Please visit \website ~to view the videos.

\section{LLM Prompts}
\label{sec:supp:prompts}

Below, we include all prompts used in our approach.
We use the same LLM pipeline and prompts in all domains and tasks.
We first outline the rationale behind our design of the LLM pipeline (\S~\ref{sec:supp:prompts:design}).
Next, we describe in detail the LLM modules and how they are used in the data generation stage (\S~\ref{sec:supp:prompts:pipeline}), summarize the general prompt structure (\S~\ref{sec:supp:prompts:structure}), and outline the API supplied to the LLM for success condition inference (\S~\ref{sec:supp:prompts:state_api}).
Finally, we show some examples of LLM completions (\S~\ref{sec:supp:prompts:completions}).

In all of our experiments, we use GPT3 (text-davinci-003) with temperature 0.0.

\subsection{LLM Pipeline Design}
\label{sec:supp:prompts:design}

Our LLM pipeline is factorized into multiple LLM modules, allowing each module's prompt to speciallize in a small reasoning skill (\eg one set of prompts for deciding whether a task involves a single or multiple objects).
We found that this not only improves the LLM's performance, but also makes designing and maintaining prompts easy.
For instance, during development, if the LLM outputs an unexpected task tree, the error could be traced back to a single module, and only that module's prompt needs to be updated.
Another convenient feature of this approach is that it also saves on token usage.
Since each module's task is small (\eg answer only ``one'' or ``multiple''), the amount of completion tokens is significantly smaller than a monolithic prompt.
Further, when a module's prompt is updated, only that module's outputs needs to be updated, allowing cost-effective approaches to cache-ing LLM's completions.

\subsection{LLM Pipeline}
\label{sec:supp:prompts:pipeline}

The recursive LLM-based planner starts with an ambiguous task description handler (Listing~\ref{list:supp:prompts:ambiguous}), which transforms ambiguous task descriptions such as ``move the block onto the catapult then shoot the block into the furthest bin'' into more specific task descriptions like ``move the block onto the catapult then shoot the block into the furthest bin by pressing the catapult's button''.
While this handler's task can occasionally overlap with the LLM planner's task, we found that it was more effective to keep them separate.

Next, given a un-ambiguous task description, the LLM planner first decides whether the planning step is necessary by checking whether the task involves touching only a single object or requiring further decomposition (Listing~\ref{list:supp:prompts:single_multiple}).
If the task involves multiple objects, it proceeds with planning (explained in the next paragraph).
If the task involves only one object part, an LLM identifies which object part name it should interact with (Listing~\ref{list:supp:prompts:obj_part}).
If the object part name is a single-link rigid object, the LLM is asked for which object it should move (the pick object part) and where (the place object part) using the prompt in Listing~\ref{list:supp:prompts:pick_place}.

In the planning step, the LLM planner outputs a list of subtasks (Listing~\ref{list:supp:prompts:planning}).
Given the recursive nature of this planning module, parent tasks also need to keep track of and propagate the current state of the environment to child tasks.
For instance, the ``open the fridge'' subtask should be followed with ``with the fridge door opened, move the eggs from the fridge ...'', such that the recursive call for moving the eggs knows it does not need to open the fridge door again.

After it has inferred the full task tree, the LLM also infers a success condition for every task in the task tree (Listing~\ref{list:supp:prompts:success}) in the form of a code-snippet.
Similar to ~\cite{codeaspolicies2022}, we inform the LLM which state API utilities are available for its usage by including import statements at the top of the file and demonstrating how they are used in the examples.

\subsection{Prompt Structure}
\label{sec:supp:prompts:structure}
All prompts start with instructions to explain to the LLM what the task is (\eg ``given an input task description, the goal is to output a list of subtasks ..''), followed by a few ``shots'' of examples, separated by a ``\#'' symbol (in text-based prompts) or a multi-line comment (in code-based prompts).
Each shot starts with a structured text encoding of the scene's object's and their parts' names in the form of a bullet list.
In the planning, success condition inference, single-or-multiple , pick-and-place, and ambiguous task description LLM tasks,
we found that it was helpful to encourage the LLM to output its reasoning (either with an explicit ``reasoning:'' field or through in-line code comments).
In contrast, we found the object part identifier task to be more effective without this explicit reasoning field.

\subsection{APIs for Success Condition Code Generation}
\label{sec:supp:prompts:state_api}

All functions take as the first argument the simulation state, which contains information on object and part names, kinematic structure, contact, all degrees of freedom, and collision meshes.

\mypara{Contact.}
This function takes as input two object (part) names, and returns whether they (or any of their parts) are in contact.

\mypara{Activation.}
A pair of functions, \texttt{check\_activated} and \texttt{check\_deactivated}, take as input an object part name and checks whether the revolute/prismatic joint connecting the object part to its parent link are near their maximum or minimum values, respectively.
This is useful for checking whether a lid is opened/closed or a button is pressed/released.

\mypara{Spatial Relations.}
We provide two spatial relations, \texttt{check\_on\_top\_of} and \texttt{check\_inside}, which takes two object (part) names and returns whether the first object (part) is on top of the second object (part) or inside the second object (part), respectively.
An object is on top of another if they are in contact and the contact normal's dot product with the up direction is greater than 0.99.
An object is inside a container if that the intersection of that object's axis-aligned bounding boxes with the container's axis-aligned bounding boxes is at least 75\% of the object's axis-aligned bounding box's volume.
This axis-aligned bounding box information can be parsed from the collision checker of most physics simulators.

\begin{lstlisting}[label={list:supp:prompts:ambiguous},language={},caption=Ambiguous task description handler's prompts]
instructions:
given an input task description, the goal is rephrase the task such that it is not ambiguous.
if the task is already specific enough, just return the original task description.
below are some examples:
#
task: stack the blocks on top of each other
scene:
 - navy block
 - maroon block
 - violet block
reasoning: the block stacking order is ambiguous. we can specify which block should be placed on which, in which order.
answer: move the maroon block onto the navy block, and the violet block on the maroon block.
#
task: move the lilac block onto the brown block
scene:
 - brown block
 - lilac block
 - yellow block
reasoning: the blocks to interact with are fully specified, so just return the original task description.
answer: move the lilac block onto the brown block.
#
task: sort the blocks based on their color's temperature onto corresponding plates
scene:
 - red block
 - orange block
 - blue block
 - purple block
 - red plate
 - blue plate
reasoning: which blocks and plates belong to the same color temperature group are ambiguous. we can specify exactly which blocks should be placed on which plate.
answer: move the red and orange blocks onto the red plate, and the purple and blue blocks onto the blue plate.
#
task: open the jar
scene:
 - jar
    + jar lid
reasoning: opening a jar is a primitive action and is fully specified, so just return the original task description.
answer: open the jar.
#
task: close the second drawer
scene:
 - drawer
    + first drawer
       + first drawer handle
    + second drawer
       + second drawer handle
    + third drawer
       + third drawer handle
reasoning: closing the second drawer is a primitive action towards a specific drawer, so just return the original task description.
answer: close the second drawer.
#
task: move the ingredients for the omelette onto the kitchen counter
scene:
 - kitchen counter
    + cupboard
       + cupboard door
          + cupboard door handle
       + salt
       + pepper
 - fridge
    + fridge door
       + fridge door handle
    + fridge top shelf
       + eggs
       + butter
       + cheese
       + milk
    + fridge bottom shelf
       + mushrooms
       + broccoli
    + freezer
       + lamb shank
       + trader joe's dumplings
       + tilapia fillet
reasoning: which ingredients belong to the omelette is ambiguous. we can specify exactly which items to take out of the fridge.
answer: move the eggs, butter, cheese, and mushrooms onto the kitchen counter and the salt and pepper onto the kitchen counter.
#
task: open the fridge, move the cheese onto the kitchen counter, and then close the fridge.
scene:
 - kitchen counter
    + cupboard
       + cupboard door
          + cupboard door handle
       + salt
       + pepper
 - fridge
    + fridge door
       + fridge door handle
    + fridge top shelf
       + eggs
       + butter
       + cheese
       + milk
    + fridge bottom shelf
       + mushrooms
       + broccoli
    + freezer
       + lamb shank
       + trader joe's dumplings
       + tilapia fillet
reasoning: which actions to perform and in which order is fully specified, so just return the original task description.
answer: open the fridge, move the cheese onto the kitchen counter, and then close the fridge.
\end{lstlisting}

\begin{lstlisting}[label={list:supp:prompts:single_multiple},language={},caption=One-or-Multiple module's prompts]
instructions:
given an input task description, the goal is to classify whether performing the task will involve touching only "one" object or "multiple" objects.
all objects start in a de-activated state (e.g., doors, drawers, cabinets, cupboards, and other objects with doors are closed, lights are off, etc.) unless specified otherwise (e.g., with the door opened).
after performing the task, objects should be reset to their de-activated state if relevant.
below are some examples:
#
task: move the blue block onto the plate
scene:
 - green block
 - blue block
 - red block
 - plate
reasoning: "moving the blue block onto the plate" involves two objects, the blue block and the plate. moving the blue block requires touching it. the plate does not have any activation state, so does not need to be touched.
answer: one.
#
task: stack the blocks on the plate
scene:
 - green block
 - plate
 - red block
 - blue block
reasoning: "stack the blocks" can be decomposed into moving the red block onto the plate, moving the green block onto the red block, and moving the blue block onto the green block. performing these steps involve touching multiple blocks.
answer: multiple.
#
task: with the red block on the plate and the orange block on the red block, move the green block onto the pink block
scene:
 - orange block
 - pink block
 - plate
 - green block
 - red block
reasoning: "moving the green block onto the pink block" involves two objects, the green block and the pink block. moving the green block requires touching it. the pink block does not have any activation state, so does not need to be touched.
answer: one.
#
task: move the lasagna into the microwave
scene:
 - microwave
    + microwave door
       + microwave door handle
 - kitchen counter
 - fridge
    + fridge door
       + fridge door handle
 - lasagna
reasoning: "moving the pasta into the microwave" involves only two objects, the lasagna and the microwave. however, it is not a primitive task because the microwave has a door (activation state), but it starts off being closed (de-activated). opening the microwave involves touching the microwave.
answer: multiple.
#
task: with the microwave opened, move the pasta into the microwave
scene:
 - microwave
    + microwave door
       + microwave door handle
 - kitchen counter
 - fridge
    + fridge door
       + fridge door handle
 - pasta
reasoning: "moving the pasta into the microwave" involves two objects, the pasta and the microwave. the microwave's door needs to be opened (activation state), but it is already opened. since the task asserts that the microwave is opened, it also does not need to be closed afterwards. this means performing the task does not involve touching the microwave.
answer: multiple.
#
task: open the microwave
scene:
 - fridge
    + fridge door
       + fridge door handle
 - dumplings
 - microwave
    + microwave door
       + microwave door handle
 - kitchen counter
reasoning: "opening the microwave" is a primitive task. it involves only one object, the microwave.
answer: one.
#
task: with the microwave opened and the sandwich in the microwave, close the microwave
scene:
 - fridge
    + fridge door
       + fridge door handle
 - sandwich
 - microwave
    + microwave door
       + microwave door handle
 - kitchen counter
reasoning: "closing the microwave" is a primitive task. it involves only one object, the microwave.
answer: one.
\end{lstlisting}
\begin{lstlisting}[label={list:supp:prompts:obj_part}, language={},caption=Object part identifier's prompts]
instructions: given an input task description, the goal is to identify which object part from the scene to interact with.

below are some examples:
#
task: stack the blue block on the plate
scene:
 - red block
 - blue block
 - green block
 - plate
answer: blue block.
#
task: with the red block on the plate, stack the green block on the red block
scene:
 - red block
 - blue block
 - green block
 - plate
answer: green block.
#
task: turn on the lights
scene:
 - light switch
 - ceiling light
 - wall
answer: light switch.
#
task: open the microwave
scene:
 - microwave
    + microwave door
       + microwave door handle
    + microwave start button
    + microwave plate
 - kitchen counter
    + cupboard
       + cupboard door
          + cupboard door handle
answer: microwave door handle.
#
task: with microwave opened and the lasagna on the kitchen counter, move the lasagna into the microwave
scene:
 - kitchen counter
    + cupboard
       + cupboard door
          + cupboard door handle
 - fridge
    + fridge door
       + fridge door handle
    + fridge top shelf
    + fridge bottom shelf
    + freezer
 - lasagna
 - microwave
    + microwave door
       + microwave door handle
    + microwave start button
    + microwave plate
answer: lasagna.
#
task: with the fridge door opened, open the cupboard
scene:
 - microwave
    + microwave door
       + microwave door handle
    + microwave start button
    + microwave plate
 - kitchen counter
    + cupboard
       + cupboard door
          + cupboard door handle
 - fridge
    + fridge door
       + fridge door handle
    + fridge top shelf
    + fridge bottom shelf
    + freezer
 - lasagna
answer: cupboard door handle.
\end{lstlisting}

\begin{lstlisting}[label={list:supp:prompts:pick_place},language={},caption=Pick \& place handler's prompts]
instructions: given an input pick and place description, the goal is to identify which object to pick and where to place among the objects listed in the scene.

below are some examples:
#
task: move the blue block on the plate
scene:
 - red block
 - blue block
 - green block
 - plate
pick: blue block.
place: plate.
#
task: with the red block on the plate, move the green block to the top of the red block
scene:
 - red block
 - blue block
 - green block
 - plate
pick: green block.
place: red block.
#
task: with microwave opened and the lasagna on the kitchen counter, move the lasagna into the microwave
scene:
 - kitchen counter
    + cupboard
       + cupboard door
          + cupboard door handle
 - fridge
    + fridge door
       + fridge door handle
    + fridge top shelf
    + fridge bottom shelf
    + freezer
 - lasagna
 - microwave
    + microwave door
       + microwave door handle
    + microwave start button
    + microwave plate
pick: lasagna.
place: microwave plate.
\end{lstlisting}

\begin{lstlisting}[label={list:supp:prompts:planning},language={},caption=Planning module's prompts]
instructions: given a input task description, the goal is to output a list of subtasks, which, when performed in sequence would solve the input task. all objects start in a de-activated state (e.g., doors, drawers, cabinets, cupboards, and other objects with doors are closed, lights are off, etc.) unless specified otherwise (e.g., with the door opened). after performing the task, objects should be reset to their de-activated state if possible. below are some examples:
#
task: move the red block onto the plate, the blue block onto the red block, and the green block on the blue block
scene:
 - red block
 - blue block
 - green block
 - plate
reasoning: no objects have activation states. the blocks can be directly placed onto the plates.
answer:
 - 1. move the red block onto the plate
 - 2. with the red block on the plate, move the blue block onto the red block
 - 3. with the red block on the plate and the blue block on the red block, move the green block onto the blue block
#
task: move the eggs, salt, and pepper onto the kitchen counter
scene:
 - kitchen counter
    + cupboard
       + cupboard door
          + cupboard door handle
       + salt
       + pepper
 - fridge
    + fridge door
       + fridge door handle
    + fridge top shelf
       + eggs
       + butter
       + cheese
       + milk
    + fridge bottom shelf
    + freezer door
       + freezer door handle
reasoning: the fridge and cupboard has doors (activation states) which start off closed (de-activated). they need to be opened before objects can be taken out of them. after the task is done, they need to be closed (reset).
answer:
 - 1. open the fridge
 - 2. with the fridge door opened, move the eggs from the fridge onto the kitchen counter
 - 3. with the eggs on the kitchen counter, close the fridge
 - 4. with the eggs on the kitchen counter, open the cupboard
 - 5. with the eggs on the kitchen counter and the cupboard door opened, move the salt onto the kitchen counter
 - 6. with the eggs and salt on the kitchen counter and the cupboard door opened, move the pepper onto the kitchen counter
 - 7. with the eggs, salt, and pepper on the kitchen counter, close the cupboard door
#
task: with the fridge door opened, move the eggs, salt, and pepper onto the kitchen counter
scene:
 - kitchen counter
    + cupboard
       + cupboard door
          + cupboard door handle
       + salt
       + pepper
 - fridge
    + fridge door
       + fridge door handle
    + fridge top shelf
       + eggs
       + butter
       + cheese
       + milk
    + fridge bottom shelf
    + freezer door
       + freezer door handle
reasoning: the fridge and cupboard has doors (activation states). the fridge's door is already opened (activated) and so don't need to be reset. the cupboard's door starts off closed (de-activated) but needs to be opened before objects can be taken out of it. after the task is done, the cupboard need to be closed (reset).
answer:
 - 1. with the fridge door opened, move the eggs from the fridge onto the kitchen counter
 - 2. with the fridge door opened and the eggs on the kitchen counter, open the cupboard
 - 3. with the fridge door opened, the eggs on the kitchen counter, and the cupboard door opened, move the salt onto the kitchen counter
 - 4. with the fridge door opened, the eggs and salt on the kitchen counter, and the cupboard door opened, move the pepper onto the kitchen counter
 - 5. with the fridge door opened, the eggs, salt, and pepper on the kitchen counter, close the cupboard door
\end{lstlisting}

\begin{lstlisting}[label={list:supp:prompts:success},language=Python,caption=Success Condition Inference module's prompts]
from utils import (
    check_contact,
    check_activated,
    check_deactivated,
    check_inside,
    check_on_top_of,
    EnvState,
)


"""
instructions:
given a input task description, the goal is to output the success condition for
that task. unless otherwise specified, all objects start in a de-activated state
(e.g., doors, drawers, cabinets, cupboards, and other containers are closed,
lights are off, etc.) unless specified otherwise (e.g., with the door opened).
after performing the task, objects should be reset to original state if possible.
"""


# robot task: touch the apple
# scene:
# - apple
#    + apple body
#    + apple stem
def touching_apple(init_state: EnvState, final_state: EnvState):
    return check_contact(
        final_state, "robotiq left finger", "apple body"
    ) and check_contact(final_state, "robotiq right finger", "apple body")


# robot task: release the cup
# scene:
# - cup
#    + cup body
#    + cup handle
def released_cup(init_state: EnvState, final_state: EnvState):
    finally_touching_cup = check_contact(
        final_state, "robotiq left finger", "cup handle"
    ) and check_contact(final_state, "robotiq right finger", "cup handle")
    finally_released_cup = (not finally_touching_cup) and (
        not final_state.gripper_command
    )
    return finally_released_cup


# robot task: move the milk carton into the shelf
# scene:
# - milk carton
# - coke can
# - shelf
def milk_carton_is_on_shelf(init_state: EnvState, final_state: EnvState):
    return check_on_top_of(final_state, "milk carton", "shelf")


# robot task: move the milk carton from the shelf
# scene:
# - milk carton
# - coke can
# - shelf
def milk_carton_is_not_on_shelf(init_state: EnvState, final_state: EnvState):
    return not check_on_top_of(final_state, "milk carton", "shelf")


# robot task: open the washing machine
# scene:
# - washing machine
#    + washing machine door
#       + washing machine door handle
#    + control panel
#       + on off button
def washing_machine_opened(init_state: EnvState, final_state: EnvState):
    return check_activated(final_state, "washing machine door")


# robot task: move the sock into the washing machine
# scene:
# - washing machine
#    + washing machine door
#       + washing machine door handle
#    + control panel
#       + on off button
# - sock
def sock_inside_washing_machine(init_state: EnvState, final_state: EnvState):
    # the washing machine can be opened (activated state) or closed (de-activated
    # state). since its activation state was not specified, the washing machine starts
    # off closed. therefore, it needs to be closed after the sock is moved inside.
    sock_inside_washing_machine = check_inside(final_state, "sock", "washing machine")
    washing_machine_door_closed = check_deactivated(final_state, "washing machine door")
    return sock_inside_washing_machine and washing_machine_door_closed


# robot task: with the washing machine opened, move the sock into the washing machine
# scene:
# - washing machine
#    + washing machine door
#       + washing machine door handle
#    + control panel
#       + on off button
# - sock
def sock_inside_washing_machine_with_washing_machine_opened(
    init_state: EnvState, final_state: EnvState
):
    # the washing machine can be opened (activated state) or closed (de-activated
    # state). the washing machine starts off opened, so it does not need to be closed
    # after the sock is moved inside.
    sock_inside_washing_machine = check_inside(final_state, "sock", "washing machine")
    return sock_inside_washing_machine
\end{lstlisting}

\subsection{Example Completions}
\label{sec:supp:prompts:completions}

Most modules are designed such that their answers are multi-choice (\eg single or multiple).
We show examples of more interesting open-ended completions for the ambiguous task description handler, planning, and success condition inference modules.
The LLM's outputs are highlighted in blue.

\subsubsection{Ambiguous Task Description Handler}
\label{sec:supp:prompts:completions:ambiguous}

\noindent\fbox{\parbox{0.97\linewidth}{\scriptsize{\texttt{{%
\query{task: move the block onto the catapult arm, then press the button to shoot the block into the closest box}\\
{\color{prompt-gray}scene:\\
- catapult\\
\hspace*{4mm} + button\\
\hspace*{4mm} + catapult arm\\
- yellow block\\
- closest box\\
- furthest box\\
- middle box\\
- table\\
reasoning:} \hlcode{which block to move onto the catapult arm is ambiguous. we can specify exactly}\\
\hlcode{which block to move onto the catapult arm.}\\
\hlcode{answer: move the yellow block onto the catapult arm, then press the button to shoot the block}
\hlcode{into the closest box.}
}}}}}\\

\noindent\fbox{\parbox{0.97\linewidth}{\scriptsize{\texttt{{%
\query{task: send the amazon package for return}\\
{\color{prompt-gray} scene:\\
- mailbox\\
\hspace*{4mm}+ mailbox lid\\
\hspace*{8mm}+ mailbox lid handle\\
\hspace*{4mm}+ mailbox flag\\
- amazon package\\
- table\\
reasoning:} \hlcode{which actions to perform and in which order is ambiguous. we can specify exactly}\\
\hlcode{ which actions to take.}\\
\hlcode{answer: open the mailbox lid, place the amazon package inside the mailbox, raise the mailbox}\\
\hlcode{flag, and then close the mailbox lid.}
}}}}}\\

\subsubsection{Planning}
\label{sec:supp:prompts:completions:planning}

\noindent\fbox{\parbox{0.97\linewidth}{\scriptsize{\texttt{{%
\query{task: move the yellow block onto the catapult arm, then press the button to shoot the block into the closest box}\\
{\color{prompt-gray} scene:\\
- catapult\\
\hspace*{4mm}+ button\\
\hspace*{4mm}+ catapult arm\\
- yellow block\\
- closest box\\
- furthest box\\
- middle box\\
- table\\
reasoning:}
\hlcode{the catapult has a button (activation state) which starts off de-activated. it }\\\hlcode{needs to be pressed to shoot the block. after the task is done, the button should be reset to}\\\hlcode{its de-activated state.
}\\\hlcode{answer:}\\\hlcode{
   - 1. move the yellow block onto the catapult arm}\\\hlcode{
   - 2. with the yellow block on the catapult arm, press the button to shoot the block into the}\\\hlcode{ closest box}\\\hlcode{
   - 3. with the yellow block in the closest box, reset the button to its de-activated state}
}}}}}\\

\noindent\fbox{\parbox{0.97\linewidth}{\scriptsize{\texttt{{%
\query{task: open the mailbox lid, place the amazon package inside the mailbox, raise the mailbox flag, and then close the mailbox lid}\\
{\color{prompt-gray} scene:\\
- mailbox\\
\hspace*{4mm}+ mailbox lid\\
\hspace*{8mm}+ mailbox lid handle\\
\hspace*{4mm}+ mailbox flag\\
- amazon package\\
- table\\
reasoning:}
\hlcode{the mailbox lid has an activation state (closed/de-activated). it needs to be}\\\hlcode{opened before the package can be placed inside. after the task is done, the lid needs to be}\\\hlcode{closed (reset).}\\\hlcode{answer:}\\\hlcode{- 1. open the mailbox lid
}\\\hlcode{- 2. with the mailbox lid opened, move the amazon package from the table into the mailbox                                                                                                                                                     }\\\hlcode{- 3. with the amazon package in the mailbox, raise the mailbox flag                                                                                                                                                                           }\\\hlcode{- 4. with the amazon package in the mailbox and the mailbox flag raised, close the mailbox lid}
}}}}}\\

\subsubsection{Success Condition Inference}
\label{sec:supp:prompts:completions:success}

\noindent\fbox{\parbox{0.97\linewidth}{\scriptsize{\texttt{{%
\query{\# robot task: move the yellow block onto the catapult arm}\\
{\color{prompt-gray}\# scene:\\
\# - catapult\\
\#\hspace*{4mm} + button\\
\#\hspace*{4mm} + catapult arm\\
\# - yellow block\\
\# - closest box\\
\# - furthest box\\
\# - middle box\\
\# - table\\
def} \hlcode{{\color{code-function}yellow_block_on_catapult_arm}(init_state: EnvState, final_state: EnvState):}\\
\hlcode{{\color{code-syntax}return} {\color{code-function}check\_on\_top\_of}(final_state, {\color{code-constant}"yellow block"}, {\color{code-constant}"catapult arm"})}
}}}}}\\

\noindent\fbox{\parbox{0.97\linewidth}{\scriptsize{\texttt{{%
\query{\# robot task: open the mailbox lid}\\
{\color{prompt-gray}\# scene:\\
\# - mailbox\\
\# \hspace*{4mm}+ mailbox lid\\
\# \hspace*{8mm}+ mailbox lid handle\\
\# \hspace*{4mm}+ mailbox flag\\
\# - amazon package\\
\# - table\\
def} \hlcode{{\color{code-function}mailbox_lid_opened}(init_state: EnvState, final_state: EnvState):}\\
\hlcode{{\color{code-syntax}return} {\color{code-function}check_activated}(final_state,  {\color{code-constant}"mailbox lid"})}
}}}}}\\

\section{Training \& Data Details.}

\subsection{Data Generation}

Our data-collection policy uses the 6DoF Exploration Primitives with the Verify \& Retry step.
For each domain, we run data generation until we get at least 500 successful trajectories per task.
Although this can be costly when tasks are long horizon with low success rates (the mailbox domain took 2 days on 256 CPU cores Intel Xeon Gold 6230R CPU @ 2.10GHz), data generation happens only once.

\subsection{Network Architecture \& Hyperparameters}

We use the same network architecture and hyperparameters for all domains.
Our task descriptions are encoded using CLIP B/32's text encoder~\cite{radford2021learning}, and projected into a 512-dimensional vector.
For each of the two camera view, we learn a separate Resnet18-based~\cite{robomimic2021} vision encoder, whose features are flattened, concatenated, and projected into a 512-dimensional vector.
The Resnet18 architecture is pre-processed by replacing BatchNorm with GroupNorm and replacing the final average pool layer with a spatial softmax pooling~\cite{robomimic2021,chi2023diffusionpolicy}.
We use an image resolution of $160\times 240$ for each view, processed with a 90\% random crop to $144\times 216$.
Finally, the proprioception is concatenated with the vision and text encoder as the condition into the diffusion policy.
We use the convolution network-based diffusion policy architecture~\cite{chi2023diffusionpolicy}.
The final network has 108 million parameters.
All networks are optimized end-to-end with the AdamW optimizer, with 5e-5 learning rate and 1e-6 weight decay, and a cosine learning rate scheduler.
For evaluation, we use an exponential moving average of all networks with a decay rate of 0.75.

\subsection{Training}

We train a separate multi-task policy for each domain using the same hyperparameters and network architecture.
For domains with only a single task, this amounts to a single-task policy.
All networks are trained for 2 days on a single NVIDIA A6000, and the best checkpoint's performance is reported.
We found that performance typically saturates around 1 day into training.

\section{Utilities Implementation}

For motion planning, we implemented rapidly-exploring random trees (RRT~\cite{lavalle1998rapidly}) with grasped-object-aware collision checking, allowing the robot to motion plan with dynamic grasping constraints.
The geometry-based grasp and placement sampler is implemented using point clouds created from depth maps, camera matrices, and segmentation maps from the simulator.
While our grasp sampler uses only geometry, kinematics, and contact information, including other grasp quality metrics (\eg stability analysis) can improve its performance.
In the placement sampler, we sample candidate place positions at points whose estimated contact normal is aligned against the gravity direction.
The revolute and prismatic joint motion primitives are implemented by checking the grasp pose relative to the joint (\eg mailbox lid handle grasp relative to the mailbox lid hinge), then performing a circular motion around the joint axis or a linear motion along the joint axis, respectively.

\begin{figure}[t]
   \centering
   \includegraphics[width=\linewidth]{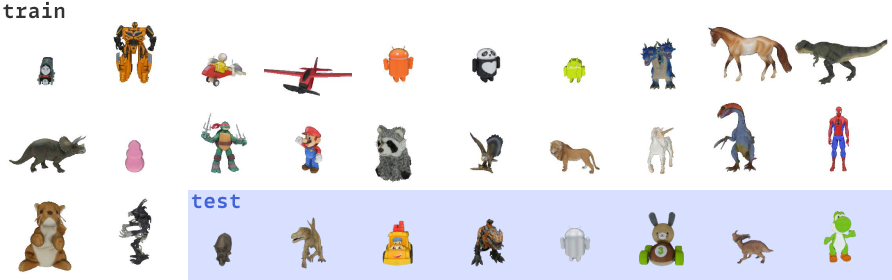}
   \caption{ \footnotesize
      \textbf{Generalization to Novel Objects.} The Transport domain requires generalization to diverse and novel object shapes and colors.
      Trained to transport 22 toys, our distilled policy generalizes to 8 novel toys (in blue section).
      All objects rendered from a fixed camera to show diversity of object size.
   }
   \label{fig:bin_transport_panel}
\end{figure}

\section{Benchmark}
\label{sec:domains_supp}

Our benchmark is built on top of the Mujoco~\cite{todorov2012mujoco} simulator, using assets from the Google Scanned Objects dataset~\cite{downs2022scannedobjects,zakka2022scannedobjectsmujoco}.
We use a table-top manipulation set-up, with a WSG50 gripper and Toyota Research Institute Finray fingers mounted on a UR5e, with a policy control rate of 4Hz.
The workspace has two cameras, one front view, which observes the entire workspace and robot, and a wrist-mounted camera, which is used to help with fine-grained manipulation~\cite{robomimic2021}.
We end episodes when any object is dropped to the floor.
Below, we clarify how we design the tasks for each domain.

\subsection{Mailbox}
\label{sec:domains_supp:mailbox}

To be considered successful, the mailbox needs to be closed with the package inside the mailbox, with the mailbox flag raised within 200 seconds (800 control cycles).
During data generation and testing, the package's planar position is uniformly random in a planar bound of dimensions [10cm, 10cm].
At evaluation, the policy has to generalize to unseen package positions.
The amazon has is a rigid object with 6DoF.
The mailbox is a fixed rigid object, with one degree of freedom for each of its revolute joints, one for the mailbox lid, and one for the mailbox flag.

\subsection{Transport}
\label{sec:domains_supp:bin_transport}

To be considered successful, the toy needs to be inside the left bin within 100 seconds.
At the beginning of each episode, a random toy 3D asset is sampled.
During data generation and testing, the toy's position is uniformly random inside the right bin, and orientation uniformly random along all three euler axes.
On top of novel randomized poses, the policy also has to generalize to unseen object instances with novel geometry.
We use 22 toys for data generation, and 8 for testing (Fig.~\ref{fig:bin_transport_panel}).
The toy is a rigid object with 6DoF, while the bins are fixed rigid objects with no DoF.
The bin asset names corresponds with their spatial location (\eg the left bin is called ``left bin'' when the scene is presented to the LLM).

\subsection{Drawer}
\label{sec:domains_supp:drawer}

This is a multi-task domain with 12 tasks, where each task involves moving one of the four objects (vitamin bottle, pencil case, crayon box, horse) into one of the three drawers (top, middle, bottom).
The task description follows the template ``move the $\langle$\textit{object}$\rangle$ into the $\langle$ \textit{drawer}$\rangle$''.
To be considered successful, the specified object needs to be inside the specified drawer within 120 seconds.
During data generation and testing, each of the four object's position is uniformly random within a planar bound of dimensions [10cm,10cm], centered around 4 evenly spaced locations along the table.
At test time, the policy has to generalize to unseen object positions in the same distribution as its data generation.

All four objects are rigid objects with 6DoF.
The drawer is a fixed articulated object with 3 DoF, one for each of the drawers.

\subsection{Catapult}
\label{sec:domains_supp:catapult}

This is a multi-task domain with 3 tasks, one for each of the three bins.
The task description follows the template ``move the block onto the catapult arm, then press the button to shoot the block into the $\langle$\textit{bin}$\rangle$'' where $\langle$\textit{bin}$\rangle$ is either closest, middle, or furthest bin.
The bin asset names corresponds with their spatial location (\eg the furthest bin is called ``furthest bin'' when the scene is presented to the LLM).

In order to be considered successful, the block needs to be inside the specified bin within 60 seconds.
This is a short amount of time, which prevents policies from retrying after failure.
The block is a rigid object with 6DoF.
The bins are fixed rigid objects with no degrees of freedom.
The catapult has two degrees of freedom, one revolute joint for the catapult arm, and one prismatic joint for the button.
This task is designed to study tool-use, and does not have any pose randomization.
Thus, different seeds affect only the policy's pseudo random samplers or the diffusion process.

We implement the catapult with a special callback function which checks whether the button sliding joint is near its max value.
If it is, then the constraint that holds the catapult arm down is disabled, releasing the spring loaded catapult arm hinge joint.

\begin{table}[t]
    \center{
        \vspace{-12mm}
        \setlength{\tabcolsep}{0.05cm}
        \begin{tabular}{lccccccccccccc}
            \toprule
            \multirow{2}{*}{Approach} & \multicolumn{3}{c}{Crayon} & \multicolumn{3}{c}{Horse} & \multicolumn{3}{c}{Pencilcase} & \multicolumn{3}{c}{Vitamin} & \multirow{2}{*}{Avg.}                                                                                                                        \\
            \cmidrule(lr){2-4}\cmidrule(lr){5-7}\cmidrule(lr){8-10}\cmidrule(lr){11-13}
                                      & B                          & M                         & T                              & B                           & M                     & T             & B             & M             & T             & B             & M             & T                    \\
            \midrule
            LLM-as-Policy (2D) 0.0    & 0.0                        & 0.0                       & 0.0                            & 0.0                         & 0.0                   & 0.0           & 0.0           & 0.0           & 0.0           & 0.0           & 0.0           & 0.0           & 0.0  \\
            (+) 6DoF Robot Utils      & 5.5                        & 0.5                       & 0.0                            & 2.0                         & 0.0                   & 0.0           & 5.0           & 0.0           & 0.0           & 2.0           & 0.0           & 0.0           & 1.3  \\
            (+) Verify \& Retry       & 48.5                       & 39.5                      & 33.0                           & 45.5                        & 32.0                  & 24.5          & 46.0          & 27.0          & 20.0          & 27.0          & 18.5          & 20.5          & 31.8
            \\\midrule
            Distill No Retry
                                      & 19.0                       & 19.0                      & 17.5                           & 13.0                        & 34.0                  & 22.5          & 27.5          & 41.0          & 39.5          & 13.5          & 12.5          & 13.5          & 22.7
            \\
            Distill  (Ours)           &
            \textbf{57.5}             & \textbf{63.0}              & \textbf{50.0}             & \textbf{62.5}                  & \textbf{59.0}               & \textbf{51.5}         & \textbf{59.5} & \textbf{72.5} & \textbf{61.5} & \textbf{46.0} & \textbf{39.5} & \textbf{46.5} & \textbf{55.8}
            \\
            \midrule
        \end{tabular}
        \caption{
            \textbf{Drawer Quantitative Results (Success Rate \%)}
            where B, M, T means bottom, middle, and top drawers.
            Averaged over 200 episodes.
        }
        \vspace{-5mm}
        \label{tab:sim_full_drawer}
    }
\end{table}

\subsection{Bus Balance}
\label{sec:domains_supp:balance}

In order to be considered successful, the bus needs to be fully balanced on top of the block within 100 seconds.
On top of testing for intuitive physics, this high precision requirement of this task was also used to test the policy's precision and ability to recover from failure, which is why we allow a generous time budget.
The task description is ``balance the bus on the block''.

The bus is a rigid object with 6DoF, dropped from a fixed location above the table with uniformly random orientation.
This means when the bus drops, it lands in different positions and orientations.
The block is fixed with no degrees of freedom.

\begin{table}[t]
    \center{
        \setlength{\tabcolsep}{0.05cm}
        \begin{tabular}{lccccccc}
            \toprule
            \multirow{2}{*}{Approach} & \multirow{2}{*}{Balance} & \multicolumn{3}{c}{Catapult} & \multicolumn{2}{c}{Transport} & \multirow{2}{*}{Mailbox}                  \\\cmidrule(lr){3-5}\cmidrule(lr){6-7}
                                      &                          & Near                         & Mid                           & Far                      & Train & Test & \\
            \midrule
            LLM-as-Policy (2D)        & 28.0
                                      & 100.0                    & 0.0                          & 0.0
                                      & --
                                      & 21.5
                                      & 0.0
            \\
            (+) 6DoF Robot Utils      & 5.5
                                      & 7.0                      & 1.0                          & 0.0
                                      & --                       & 35.0
                                      & 0.0
            \\
            (+) Verify \& Retry       & 45.0
                                      & 16.3                     & 3.3                          & 2.2
                                      & --                       & \textbf{82.0}
                                      & 3.0
            \\\midrule
            Distill No Retry          & 67.5
                                      & 2.5
                                      & \textbf{56.5}            & \textbf{56.5}

                                      & 31.0
                                      & 32.5
                                      & 0.0
            \\
            Distill (Ours)            & \textbf{79.0}
                                      & \textbf{78.0}            & 52.0                         & 45.0
                                      & \textbf{74.0}            & 80.0
                                      & \textbf{62.0}
            \\
            \midrule
        \end{tabular}
        \caption{
            \textbf{Full Quantitative Results (Success Rate \%).
            } Averaged over 200 episodes.
        }
        \label{tab:sim_full}
    }
\end{table}

\section{Full Results}
We include the full results for all tasks in the drawer domain in Table~\ref{tab:sim_full_drawer}, and all other domains in Table~\ref{tab:sim_full}.
We omit data generation baseline numbers on the train set in the transport domain, since they are non-learning approaches.
All approaches are evaluated on 200 different seeds, which controls pose randomization, which asset is sampled, the pseudo-random robotic utility samplers, and the pseudo-random diffusion process.
We make one exception in the catapult domain, where due to the low success rates of getting the block into the middle and far bin, we run evaluation until there are 500 successful trajectories per task, then report the average success rate.
Since the time limit for the catapult is short, the data-collection policy will not have enough time to retry, leading to identical numbers with the baseline data-collection policy without verify \& retry.

In the drawer domain, we observe that the task is more difficult for:
\begin{enumerate}
   \item \textbf{Larger objects:}
         The most challenging objects are the vitamin bottle and the horse toy, both of which are too large to fit the drawer if they are in an upright orientation.
         This means to be effective at this task, the robot should perform sideway grasps on these objects, such that downstream placement is easier.
         In contrast, the small crayon box is has the highest success rates amongst the data-collection policies.
   \item \textbf{Top drawer:}
         We observe interacting with this drawer often brings the robot close to its kinematic reach range.
         This means slight imprecision in the policy's predicted actions or small shifts in the grasped object (which is unaccounted for during motion planning) in execution could lead to failure.
         For instance, while moving the objects inside the top drawer, the grasped object could collide with the drawer, causing the grasped object to drop or the drawer to close.
   \item \textbf{Planar Action Primitives:}
         A top-down grasp on the drawer handle will typically be in collision with the drawer's body.
         Thus, in LLM-as-Policy (2D)'s first action to open the drawer, its call to the motion planner will fail due to an invalid goal configuration.
\end{enumerate}

\begin{figure}[t]
   \centering
   \includegraphics[width=\linewidth]{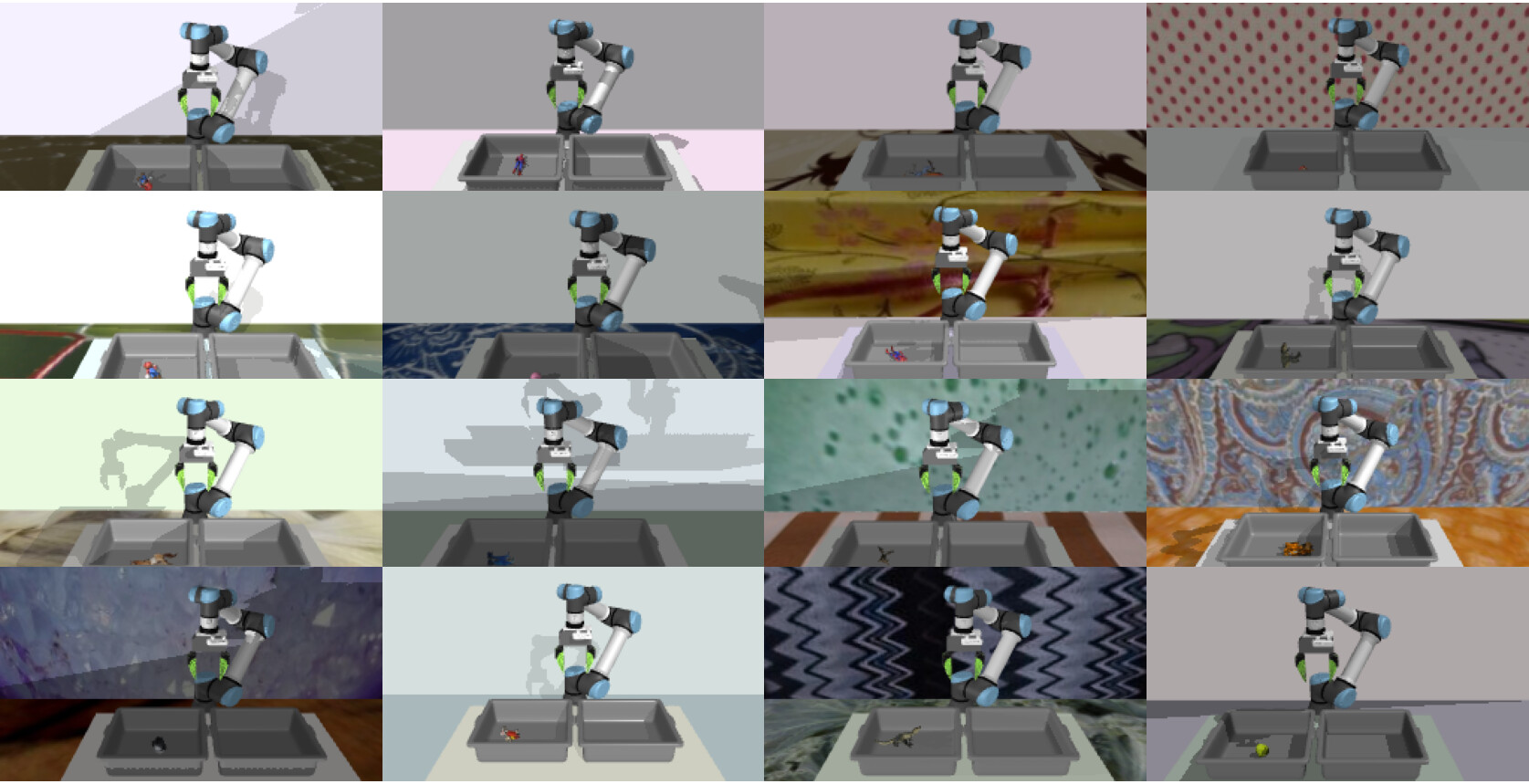}
   \caption{ \footnotesize
      \textbf{Domain Randomization.}
      To facilitate Sim2Real transfer, we train our policy on lighting, texture, and camera pose randomized scenes.
   }
   \label{fig:domain_rand}
\end{figure}

\begin{wrapfigure}{l}{0.48\textwidth}
   \centering
   \vspace{-5mm}
   \includegraphics[width=\linewidth]{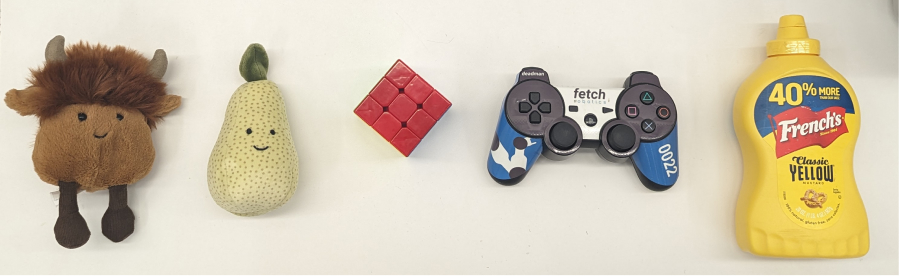}
   \caption{ \footnotesize \textbf{Real World Objects}.
   }
   \label{fig:real_world:objs}
   \vspace{-2mm}
\end{wrapfigure}

\section{Real World Evaluation}

We train a separate policy for real-world transfer on domain randomized scenes (Fig.~\ref{fig:domain_rand}).
We evaluate our policy on a real UR5e robot with a WSG50 gripper and Toyota Research Institute Finray fingers, matching our simulation set-up.
We use five unseen objects (Fig.~\ref{fig:real_world:objs}), ranging in shape, size, and visual appearance.
Each object is evaluated on 10 episodes, with the object placed at a random pose on the right bin.
We observe 70\%, 80\%, 60\%, 80\%, and 90\% for the pear, monster, rubiks cube, fetch controller, and mustard bottle respectively, giving a mean success rate of 76\%.

\end{appendix}

\end{document}